\documentclass[manuscript, nonacm=true]{acmart}

\AtBeginDocument{\providecommand\BibTeX{{\normalfont B\kern-0.5em{\scshape i\kern-0.25em b}\kern-0.8em\TeX}}}

\usepackage{algorithm2e}
\usepackage{graphicx}
\usepackage{textcomp}
\usepackage{xcolor}
\usepackage{url}
\def\BibTeX{{\rm B\kern-.05em{\sc i\kern-.025em b}\kern-.08em
    T\kern-.1667em\lower.7ex\hbox{E}\kern-.125emX}}

\usepackage{algpseudocode}

\usepackage[acronym]{glossaries}
\usepackage{xspace}
\graphicspath{{multimedia/}}

\begin{document}

\title[ChatGPT and Other LLMs as Evolutionary Engines for Online Interactive Collaborative Game Design] {ChatGPT and Other Large Language Models as Evolutionary Engines for Online Interactive Collaborative Game Design}

\author{Pier Luca Lanzi}
\email{pierluca.lanzi@polimi.it}
\orcid{0000-0002-1933-7717}
\affiliation{\institution{Politecnico di Milano}
\city{Milano}
\country{Italy}
}

\author{Daniele Loiacono}
\email{daniele.loiacono@polimi.it}
\orcid{0000-0002-5355-0634}
\affiliation{\institution{Politecnico di Milano}
\city{Milano}
\country{Italy}
}

\begin{abstract}
Large language models (LLMs) have taken the scientific world by storm, changing the landscape of natural language processing and human-computer interaction. These powerful tools can answer complex questions and, surprisingly, perform challenging creative tasks (e.g., generate code and applications to solve problems, write stories, pieces of music, etc.). 
In this paper, we present a collaborative game design framework that combines interactive evolution and large language models to simulate the typical human design process. We use the former to exploit users' feedback for selecting the most promising ideas and large language models for a very complex creative task---the recombination and variation of ideas. In our framework, the process starts with a brief and a set of candidate designs, either generated using a language model or proposed by the users. Next, users collaborate on the design process by providing feedback to an interactive genetic algorithm that selects, recombines, and mutates the most promising designs.
We evaluated our framework on three game design tasks with human designers who collaborated remotely.
 \end{abstract}




\maketitle

\eject\vfill
\section{Introduction}
Design is a collaborative problem-solving activity that begins with a design brief providing an approximate description of a target objective. In the opening (divergent) phase, several initial ideas of possible solutions are proposed. These are the seeds for the next exploration (emergent) phase in which ideas are modified and combined to create new solutions over a series of iterations. Finally, in the closing (convergent) phase, developed ideas are critically examined, assessed, and the process result is synthesized \cite{norman:design,gray:2010:gamestorming}. 
Interactive evolution \cite{949485} can model collaborative design processes effectively, and it has been widely applied to design and creative activities like fashion \cite{kim2000application,Kim2005,10.1007/978-3-319-55750-2_20,MOK20131442}, 
music composition \cite{xu2010emotional,DBLP:conf/nime/BrownHEGFM18,DBLP:journals/connection/HooverS09,DBLP:conf/gecco/HooverSS11},
video games \cite{hastings2010,DBLP:conf/aiide/RisiLDHS12}
and art in general \cite{DBLP:conf/ieeealife/KowaliwMD11,DBLP:journals/tec/KowaliwDM12}.
However, its application has been limited to domains where admissible design solutions could be encoded, randomly generated, mutated, and recombined. For example, it has never been applied to evolve free-form texts that would need advanced operators to maintain semantic coherence in  generating the results.  

Large Language Models (LLMs) have changed the landscape of natural language processing and human-computer interaction. These powerful tools can answer complex questions and perform challenging creative tasks such as 
writing screenplays and theater scripts \cite{https://doi.org/10.48550/arxiv.2209.14958}, 
solving complex math problems \cite{https://doi.org/10.48550/arxiv.2110.14168},
theorem proofing \cite{https://doi.org/10.48550/arxiv.2009.03393},
generating code to solve specific tasks \cite{https://doi.org/10.48550/arxiv.2107.03374}, generate descriptions from images \cite{DBLP:conf/icml/ChenRC0JLS20}, etc. 
More interestingly, they can also generate random texts from high-level instructions, combine two or more texts, and create variations of existing texts---all while maintaining semantic coherence in the generated texts. These functionalities can easily map to the genetic operators at the core of most evolutionary algorithms. Thus, we can use LLMs to implement evolutionary operators that can manage free-form text individuals while also maintaining semantic coherence---something that would be difficult, if not impossible, to achieve with traditional representations. 

In this paper, we present an online collaborative design framework that combines interactive evolution with large language models for evolving game design ideas (\textit{game concepts}) represented as free-form texts. The online interactive evolutionary algorithm shows the ideas to the human designers and collects their feedback; the large language model implements all the operators for the random generation, the recombination (crossover), and the variation (mutation) of ideas. Our framework is very general in that it can support any design process based on textual information (and could be easily extended to integrate visuals integrating tools like Dall-E 2 \cite{dall-e-2}); it can use any viable LLM and therefore, it is intrinsically multilingual; it can be accessed using any medium that can visualize texts like a webpage  \cite{DBLP:conf/gecco/CardamoneLL11,DBLP:journals/asc/CardamoneLL15}, a 2D-grid displayed on screen \cite{DBLP:conf/evoW/BontragerLTR18,10.1145/2576768.2598353}, or any viable messaging platform that supports user polls or other ways to collect users’ feedback. 

We performed a preliminary evaluation of our framework on three creative tasks, the design of a board game, the design of a video game, and the generation of game ideas during the 2023 Global Game Jam. The evaluation involved around 80 users (junior and senior game designers). 
The goal of the two game design tasks was to simulate a typical idea-generation process; accordingly, we set a fixed time limit of four working days for each task. The interactive evolutionary algorithm communicated with the human designers using the Telegram messaging platform \cite{telegram} while the genetic operators were implemented using ChatGPT \cite{ChatGPT_2021}. Each task was initiated with a design brief \cite{gray:2010:gamestorming} that was sent to the designers as the first message on a Telegram group. The brief was followed by a series of initial design ideas (some randomly generated using the LLM, some collected from existing designs) --- the opening (divergent) phase \cite{gray:2010:gamestorming}. Then, over the next four days, users evaluated the design ideas they received using the Telegram poll system; the evolutionary algorithm employed the users’ feedback to select, recombine and mutate the existing ideas into new ones presented next. 
Instead, the evaluation performed during the 2023 Global Game Jam was conceived as an open-ended brainstorming activity that lasted less than 24 hours to fit the tight schedule of the event.

There are few examples of applications of LLMs to idea generation and creativity in games. For example, Zhu and Luo \cite{https://doi.org/10.48550/arxiv.2204.09658} explored a generative approach for design ideation combining a language model and a knowledge based of existing patents to generate new ideas. van Stegeren and Myundefinedliwiec \cite{10.1145/3472538.3472595}
investigated the application of GPT-2 for generating dialogues in role-playing games for non-player characters while 
Värtinen et al. \cite{9980408} applied GPT-2 \cite{gpt2} and GPT-3 \cite{gpt3} to generate description of quests in the same type of games. Frans \cite{9619126} presented a framework for building \emph{Language Model Games} whose main mechanic involved players manipulating a language model into behaving in a desired manner with a demo game called AI Charades. However, to the best of our knowledge, our framework is the first example of interactive evolutionary algorithm that apply large language models to implement genetic operator and combines them with interactive evolution to mimic a full-fledged design session from start to end. 

\section{Related Work}
\label{sec:related_chatgpt}
Interactive evolution \cite{949485} dates back to Richard Dawkins's  
	\textit{The Blind Watchmaker} and \textit{Biomorphs} \cite{dawkins:blind_watchmaker}.
Over the years, this approach has been applied to several creative tasks, for which human evaluation might provide a better (if not the only) 
alternative to the definition of an objective (or fitness) function to evaluate candidate solution, such as 
fashion \cite{kim2000application,Kim2005,10.1007/978-3-319-55750-2_20,MOK20131442}, 
ergonomics \cite{brintrup2008ergonomic,WANG2020102901},
yacht design \cite{KHAN2019106462},
visual effects \cite{DBLP:conf/eurogp/EbnerRA05,pcgfragment},
synthetic images \cite{DBLP:conf/chi/SecretanB08}, 
music \cite{xu2010emotional,DBLP:conf/nime/BrownHEGFM18,DBLP:journals/connection/HooverS09,DBLP:conf/gecco/HooverSS11},
video games \cite{hastings2010,DBLP:conf/aiide/RisiLDHS12}
and art in general \cite{DBLP:conf/ieeealife/KowaliwMD11,DBLP:journals/tec/KowaliwDM12}.

Interactive evolutionary frameworks collect human evaluations either implicitly or explicitly. In the former case, the systems track human actions, sometimes building a model of users' behavior and deriving an assessment of their preferences \cite{5740836,DBLP:conf/gecco/HornbyB12}. Galactic Arms Race (GAR) \cite{hastings2010} is an example of implicit evaluation. It was the first video game using interactive evolution for content generation; it included a weapon system that automatically evolved new weapons based on players' behavior.
Petalz \cite{DBLP:conf/aiide/RisiLDHS12} was a Facebook game about growing a garden of procedurally generated flowers;
players would select what flowers they liked most and wanted to breed; human evaluation used by the interactive evolutionary algorithm 
was implicitly implemented within the farming mechanics. 

Most often, interactive evolutionary frameworks involve an explicit human-based evaluation of candidates and, similarly to Dawkins' \textit{Biomorphs} \cite{dawkins:blind_watchmaker}, have a grid-like interface that allows users to evaluate candidate designs iteratively. 
Hoover et al. \cite{DBLP:journals/connection/HooverS09,DBLP:conf/gecco/HooverSS11}
	integrated interactive evolution with functional scaffolding for generating drum tracks \cite{DBLP:journals/connection/HooverS09} 
	and accompaniments for existing musical pieces \cite{DBLP:conf/gecco/HooverSS11} using a compositional pattern-producing network.
	At first, users select a musical piece (in MIDI format) and which parts should be elaborated by the evolutionary algorithm (e.g., the piano, guitar, or bass guitar). 
	Next, users customize and refine the computer-generated accompaniment through an interactive selection and mutation of compositional networks. 
    Noticeably, the user interface in \cite{DBLP:conf/gecco/HooverSS11} is 
    similar to other music composition tools instead of the grid-based structure used in most frameworks.
Cardamone et al. \cite{DBLP:conf/gecco/CardamoneLL11,DBLP:journals/asc/CardamoneLL15} developed a tool that combined 
	an interactive genetic algorithm with procedural content generation to evolve tracks for two open-source racing games.
	The tool had a web frontend that remained active for several years and currently appears to be discontinued, although still online.
Ølsted et al. \cite{7257069}
	developed a tool for the interactive evolution of first-person-shooter maps inspired by the popular multiplayer game Counter-Strike.
While the tool of \cite{DBLP:conf/gecco/CardamoneLL11,DBLP:journals/asc/CardamoneLL15} is external to the actual racing games, 
	the tool of Ølsted et al. \cite{7257069} is integrated within the game (FPSEvolver) and operated \textit{live},
	collecting players' feedback during the usual map voting and selection process---a typical phase in multiplayer first-person-shooter games. 
Pirovano et al. \cite{DBLP:conf/gamesem/PirovanoML15}
	developed a tool for creating 3D sword models that combined procedural content generation, interactive evolution, and a grid-based 
	interface to let users explore such a peculiar design space. 
Generative Adversarial Networks (GANs) \cite{DBLP:journals/cacm/GoodfellowPMXWO20} represented a milestone in procedural content generation and have been applied to various creative tasks. They learn the probability distribution that generates a collection of training examples and can generate more samples from randomly generated input vectors of Gaussian noise, which form the latent space. 
Several authors developed methods to explore the latent space to guide the generation of examples \cite{DBLP:conf/gecco/VolzSLLSR18,DBLP:conf/cig/GiacomelloLL19},
	including interactive evolution \cite{DBLP:conf/evoW/BontragerLTR18,10.1145/3377930.3389821}.
Bontrager et al. \cite{DBLP:conf/evoW/BontragerLTR18} combined 
generative adversarial networks for image generation with interactive evolution to explore the latent space to improve 
	the quality of sampled images for specific targets.
Schrum et al. \cite{10.1145/3377930.3389821} applied interactive evolution for  exploring the latent space for generating levels for the video games \textit{Super Mario Bros.} and \textit{The Legend of Zelda}. 
Both \cite{DBLP:conf/evoW/BontragerLTR18} and \cite{10.1145/3377930.3389821}
	require an explicit human-based evaluation using a grid-like interface similar to the ones used in  \cite{dawkins:blind_watchmaker,DBLP:conf/gecco/CardamoneLL11,DBLP:journals/asc/CardamoneLL15,DBLP:conf/gamesem/PirovanoML15}.

\section{The Framework Architecture}
\label{sec:architecture}
\begin{figure}[t]
    \centering
    \includegraphics[width=\columnwidth]{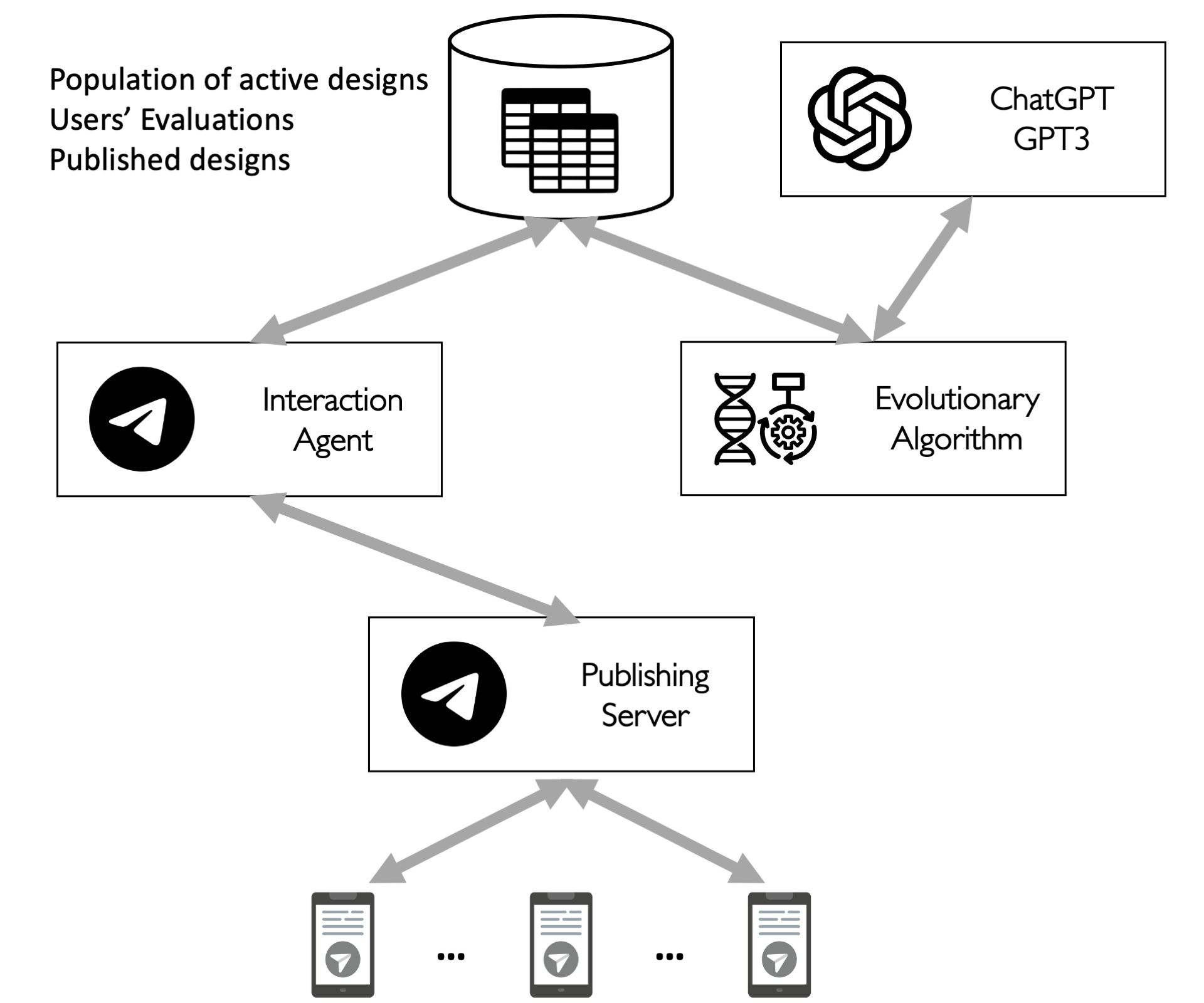}
    \caption{The framework architecture when collaboration is based on the Telegram messaging platform.}
    \label{fig:framework}
\end{figure}

Our framework combines interactive evolution and large language models to simulate the typical human game design process. It comprises three components: (i) the online database of game design ideas; (ii) the evolutionary algorithm; and (iii) the agent responsible for publishing the designs and collecting users' feedback. Figure~\ref{fig:framework} shows the structure of our framework; in this case, the interaction with the human designers uses the Telegram messaging system, which is the one we employed in the experiments discussed in this study. 

\medskip\noindent\textbf{The database} contains (i) a list of game design ideas that have been published and are currently under evaluation; (ii) the list of submitted evaluations that are used to compute the fitness; (iii) the current population of active design ideas; this includes both the designs that been already published and under evaluation by the users and the ones that have been already generated but not published yet.

\medskip\noindent\textbf{The evolutionary engine} is implemented as a steady-state genetic algorithm that, at each iteration, (i) selects promising game design ideas from the current population; (ii) applies recombination and mutation (implemented using a large language model like ChatGPT or another LLM like DaVinci GPT-3 \cite{gpt3}); (iii) inserts the newly generated individuals in the population; finally, (iv) deletes individuals to keep the population size constant. 

\medskip\noindent\textbf{The interaction agent} manages the communication between the framework and the publishing server; it publishes the design concepts in the population and collects the users' feedback. In Figure~\ref{fig:framework}, we show the architecture used in the experiments discussed in this paper with an interaction agent implemented as a Telegram bot that publishes the design concepts on a Telegram group through the Telegram servers. However, our framework also supports publishing using web servers (as done in \cite{DBLP:conf/chi/SecretanB08,DBLP:conf/gecco/CardamoneLL11}). 

Note that, when using a publishing-subscribe platform like Telegram, the publication of design concepts is timed to limit the number of concepts (and message notifications) the users receive at any given time (see Section~\ref{sec:evaluation}). Accordingly, we maintain two separate tables, one for the population of active designs (Figure~\ref{fig:framework}) and one for the published designs currently being evaluated. In contrast, when interacting through a web server, the agent will publish the population all at once; users will autonomously decide how many concepts they want to view and evaluate using the online navigation interface (similarly to \cite{DBLP:conf/chi/SecretanB08,DBLP:conf/gecco/CardamoneLL11}).

\medskip\noindent\textbf{User Evaluation.} 
Most of the interactive evolutionary algorithms we examined (Section~\ref{sec:related_chatgpt}) ask for qualitative evaluations based, for example, on basic voting systems \cite{7257069,10.1145/3377930.3389821}, scales with a limited number of values  \cite{DBLP:conf/gecco/CardamoneLL11,DBLP:journals/asc/CardamoneLL15}, or let users actively select what individual should go to the next generation \cite{10.1145/2576768.2598353,DBLP:conf/evoW/BontragerLTR18}. In our framework, we adopted a simple three-value scale (positive, neutral, and negative) that let users express how they felt about a given design idea. Note that, the evaluation is anonymous and users cannot see how other people evaluated an idea before they submitted their evaluation. When using the Telegram poll-based interface, only the summary statistics about the submitted evaluations will appear on the group chat; when using the web-based interface, users can see only their evaluation. 

\subsection{The Workflow}
The process starts with a short \textit{design brief}, describing the activity goals \cite{wiki:design_brief}. The evolutionary algorithm uses the brief to generate the initial population either (i) by asking ChatGPT or another language model to generate random solutions (similarly to the traditional population initialization in evolutionary algorithms); (ii) by introducing human-designed concepts originating from the brief (e.g., by asking participants to propose solutions or taking existing well-known solutions); or (iii) by combining the previous options (similarly to the half and half initialization in Genetic Programming \cite{koza:1992}). The individuals in the initial population are published either at once (when collaboration is web-based) or timely (when using a publishing-subscribe platform like Telegram). Next, the framework waits for feedback from the users, who can score each published concept. When sufficient evaluations have been submitted for all the published concepts, the evolutionary algorithm starts its select-recombine-mutate cycle: it selects two individuals from the population using tournament selection of size two; with a given probability, it applies recombination to generate one offspring that is then mutated; finally, the offspring is added to the population while the worst individual is deleted to keep the population size constant. The new individual becomes eligible for publication; depending on the platform used, the individual is published immediately or in a given time frame. The next iteration of the evolutionary algorithm will start when a sufficient number of evaluations have been submitted for the newly published individuals. 
 
\section{Implementing Genetic Operators with Large Language Models}
\label{sec:evolutionary}
LLMs are queried using questions (i.e., \textit{prompts}) expressed in natural language. The quality of their answers heavily relies on how well the prompts can specify the objective for the target LLM. In our framework, individuals represent game design ideas expressed as free-form text; all the operators manipulating them are implemented by querying an LLM. Accordingly, we initially studied how to design prompts so that the LLM answers could mimic the behavior of the three operators we needed to implement the evolutionary algorithm: random initialization of individuals, recombination or crossover, and variation or mutation.

\subsection{Random Initialization of Individual}
In our framework, the population can be initialized (i) with random individuals, (ii) with existing designs, or (iii) with a combination of the previous approaches. 
Random individuals (game ideas) ignite the design process and need to provide an approximate indication of the target objective.
In our initial analysis of feasible prompts to implement evolutionary operators we noted that it is convenient to ask the system to structure the answer. 
Accordingly, the prompts used to generate the initial random game ideas consist of (i) a design brief defining the objective and its constraints in an organized structure; and (ii) a request to \emph{act as a game designer} and generate a novel design based on the brief.	
Table~\ref{tab:prompts}a shows a prompt to generate a random idea for a generic board game. The bold text shows the actual request while the remaining part specifies how the answer should be structured. Note that, the prompt also specifies the length of each section to avoid very long game ideas that could be more difficult to evaluate by the designers. 
When the request is less generic, the prompt begins with a text that serves as inspiration or as an example of the objective.
Table~\ref{tab:prompts}b shows a prompt to generate a video game that uses a minimal amount of resources. 
This example is inspired to the study of minimalism in video game design  \cite{DBLP:conf/digra/Myers09,DBLP:conf/fdg/NealenSB11}.
The first sentence of the prompt is a statement that serves as an inspiration and sets the basis of what is going to be asked. The second sentence (in bold) shows the actual request (as in the previous case). The remaining text specifies the answer structure.
Table~\ref{tab:prompts}c shows an example of prompt that was used to generate random individuals for an experiment run during the 2023 Global Game Jam\footnote{\url{https://globalgamejam.org}} in which the goal was to create a game based on the theme \emph{roots}. Also in this case, the first sentence was used as a sort of inspiration. However, since there could be many interpretations of \emph{roots} (it could refer to cultural heritage, botanic, etc.), the sentence contained the marker $\langle INTERPRETATION \rangle$ which was replaced with sentences suggesting different  interpretations (Table~\ref{tab:prompts}d). The process generated a great variety of prompts for the same theme.

\begin{table*}
	\begin{footnotesize}
	\begin{tabular}{c}\hline
		\\
		\parbox{.9\textwidth}{
			\textbf{Act as a board game designer and create a unique board game with the following restrictions:} 
			the game can involve only a board (of any layot) and a set of pieces (of any number, and type). Organize your response as follows: 1) Name of the Game; 2) Number of players and cooperative/competitive game; 3) Game Board: describe the layout of the game board, the number and shape of its tiles (max 200 characters); 4) Game Pieces: describe the number of pieces, their type, and their distribution among players (max 300 characters); 5) Rules: provide a detailed explanation of how the game is played, including how to move pieces and take turns (max 800 characters); 6) Objective: provide a clear description of the win and lose conditions of the game (max 200 characters). Keep the game description as simple as possible, without including visual details, theme and story.
		} \\
		\\
		\hline
		\\
		(a) \\		
		\\\hline
		\\
		\parbox{.9\textwidth}{
			A white pixel is the minimum amount of information we can show on-screen, and pressing a key (or a button) is the least interaction we can ask players. 
			\textbf{Act as a game designer and design a minimalist game. } Organize your response as follows: 1) the name of the game; 2) game concept:  describe the idea, the core mechanics, the goal of the game, and player controls (max 1000 chars); 3) game resources: describe the graphical elements involved by the game, which type of player input is required, and any additional resources required by the game (e.g., sound); 4) level design: describe how to do the level design for this game (max 300 chars); 5) game instructions: provide the game instructions for the player (max 300 chars). Do not add additional information.

		}\\
		\\\hline
		\\
		(b)\\
		\\\hline
		\\
		\parbox{.9\textwidth}{
			\textbf{Act as a game designer and describe (max 1000 chars), a concept for a video game about the theme} "roots: $\langle INTERPRETATION \rangle$". 			
			Both the game mechanics and the interaction of the players with the game should involve elements directly connected to the theme of the game.

			Organize your response as follows: 1) name of the game; 2) game concept:  describe the idea, the core mechanics, the goal of the game, and player controls (max 800 chars); 3) level design: describe how to do the level design for this game (max 400 chars); 4) game instructions: provide the game instructions for the player (max 300 chars). Do not add additional information.
		}\\
		\\\hline
		\\
		(c)\\
		\\\hline
		\\
		\parbox{.9\textwidth}{
			\begin{itemize}
				\item "the part of a plant that is below the ground and that absorbs water and minerals from the soil"
				\item "the part of a tooth within the socket"
				\item "one or more progenitors of a group of descendants"
				\item "the essential core"
				\item ... 
			\end{itemize}
		}\\
		\\\hline
		\\
		(d)\\
		\\
		\\\hline
		\\
		\parbox{.9\textwidth}{
			$\langle BRIEF \rangle$. 
			Given these two examples of minimalist game.
			
			EXAMPLE1: $\langle INDIVIDUAL \rangle$

			EXAMPLE2: $\langle INDIVIDUAL \rangle$
			
			\textbf{Act as a game designer and recombine these two games to create a novel game.} 
			In your response do not include any introduction or final comment. In the response, please avoid references to the two games recombined.			
		}\\
		\\\hline
		\\
		(e)\\
		\\
		\\\hline
		\\
		\parbox{.9\textwidth}{
			$\langle BRIEF \rangle$. Given the following example of minimalist game.
	
			EXAMPLE: $\langle INDIVIDUAL \rangle$
	
			\textbf{Act as a game designer and design a novel game as a variation of the given example, by changing only} $\langle MUTATION \rangle$. 
			In your response do not include any introduction or final comment. In the response, please avoid explicit reference to the original game.
		}\\
		\\\hline
		\\
		(f)\\
		\\		
	\end{tabular}
	\end{footnotesize}

	\caption{Example of prompts for generating (a) a generic board game; (b) a minimalistic video game; (c) a game based on the theme \textit{roots} in which the marker $\langle INTERPRETATION \rangle$ can be replaced with sentences suggesting different theme interpretation, like the ones listed in (d). Prompts to implement (e) recombination (crossover) and (f) variation (mutation); $\langle BRIEF \rangle$ identify the brief used to generate the random individuals, $\langle INDIVIDUAL \rangle$ an individual selected from the population, and $\langle MUTATION \rangle$ a high-level characteristic of gameplay.}
	\label{tab:prompts}
\end{table*}

\subsection{Recombination (Crossover)}
The prompt to implement the recombination of two individuals has a general structure (Table~\ref{tab:prompts}e). It starts with the same brief used to generate random individuals. For example, when recombining two video games using minimal resources, the marker $\langle BRIEF \rangle$ would be replaced with the sentence ``A white pixel is the minimum amount [...]'' from Table~\ref{tab:prompts}b. The brief is followed by two examples of games in which the markers $\langle INDIVIDUAL \rangle$ are replaced with the text description of the two parents selected for recombination. 
Next, the actual request (highlighted in bold) and two sentences specifying a series of constraints to avoid answers that are blunt rephrasing of the parents. Note that, the recombination prompt generates one single offspring and to generate a second offspring (as in traditional crossover) another prompt using two different parents should be used. We tried several prompts to generate two offspring at once but we noted that such approach would lead with very similar offspring so we opted for a recombination prompt that would generate only one offspring.

\subsection{Variation (Mutation)}
The prompt for variation also starts with the brief used to generate random individuals (marker $\langle BRIEF \rangle$ in Table~\ref{tab:prompts}f) followed by (i) an existing game idea to be mutated (marker $\langle INDIVIDUAL \rangle$), (ii) the request  to the LLM (highlighted in bold) specifying the focus of the mutation (marker $\langle MUTATION \rangle$), and (iv) two  sentences specifying a series of constraints to avoid answers that would be simple rephrasing of the parent. Marker $\langle MUTATION \rangle$ specifies the focus of the mutation and it is randomly replaced with a gameplay aspects such as,  \textit{the goal of the game}, \textit{the game's resources}, \textit{the level design}, \textit{the player's input}, etc. We experimented with mutation operators that would work at a lower level of detail (e.g., by substituting some keywords of the game idea with synonyms or antonyms generated using word embeddings); however, the approach proved less effective as it did not introduce enough variation. 
 
\section{Experimental Evaluation}
\label{sec:evaluation}
We performed a preliminary evaluation of our framework with human subjects (junior and senior game designers) on three design tasks. One focused on designing a video game with the least amount of resources possible while having interesting mechanics---a minimalistic video game \cite{DBLP:conf/digra/Myers09,DBLP:conf/fdg/NealenSB11}. One concerned the creation of a generic board game using only one board of any layout and a set of pieces. Finally, we evaluated our framework with people participating in the 2023 Global Game Jam as a tool for the initial brainstorming phase of the jam. All the experiments were run using the Telegram-based version of our framework (Figure~\ref{fig:framework}). 

\subsection{Board Game and Minimalistic Video Game Design}
These two experiments were aimed at simulating the typical idea-generation process in a controlled environment (e.g., a design studio) that usually is well-structured \cite{gray:2010:gamestorming}, has clear goals, and has a fixed duration. We set up two Telegram groups, one for each task. 
We sent a call for volunteers to game designer communities on Facebook and mailing lists, providing a link for joining each Telegram group. 
In total, 44 individuals joined one or both groups, with 35 actively engaging in the experiments. Among these participants, 23 were involved in both experiments, while 8 participated exclusively in the video game task, and 4 in the board game task.
Not all participants disclosed their identities, limiting our ability to provide comprehensive demographic data. In response to concerns regarding the availability for continuous feedback throughout the day, we scheduled the publication of design ideas during three distinct periods: morning, noon, and late afternoon. The experiments began with a brief that we posted on each Telegram group with information about the timing of the postings so that people could plan their participation.
The participants were not informed about the evolutionary algorithm that was running in the background and about the use of large language models. Instead, we informed them that AI will be used to support the design process based on their feedback.
The experiments lasted four days, Tuesday to Friday; we allowed a grace period of a few hours, until Saturday morning, to let participants submit their final evaluations. We wanted to focus the design process and limit the number of ideas that the participants had to evaluate initially (Section~\ref{sec:architecture}). Accordingly, we chose a small population size (10 individuals); selection, recombination, and variation were implemented using ChatGPT \cite{ChatGPT_2021}; they were triggered as soon as 25 new evaluations were submitted; selection was implemented using a tournament selection of size 2; recombination probability was set to 0.7; variation was always applied. 

\begin{figure}
	
	\centering
	\includegraphics[width=.8\columnwidth]{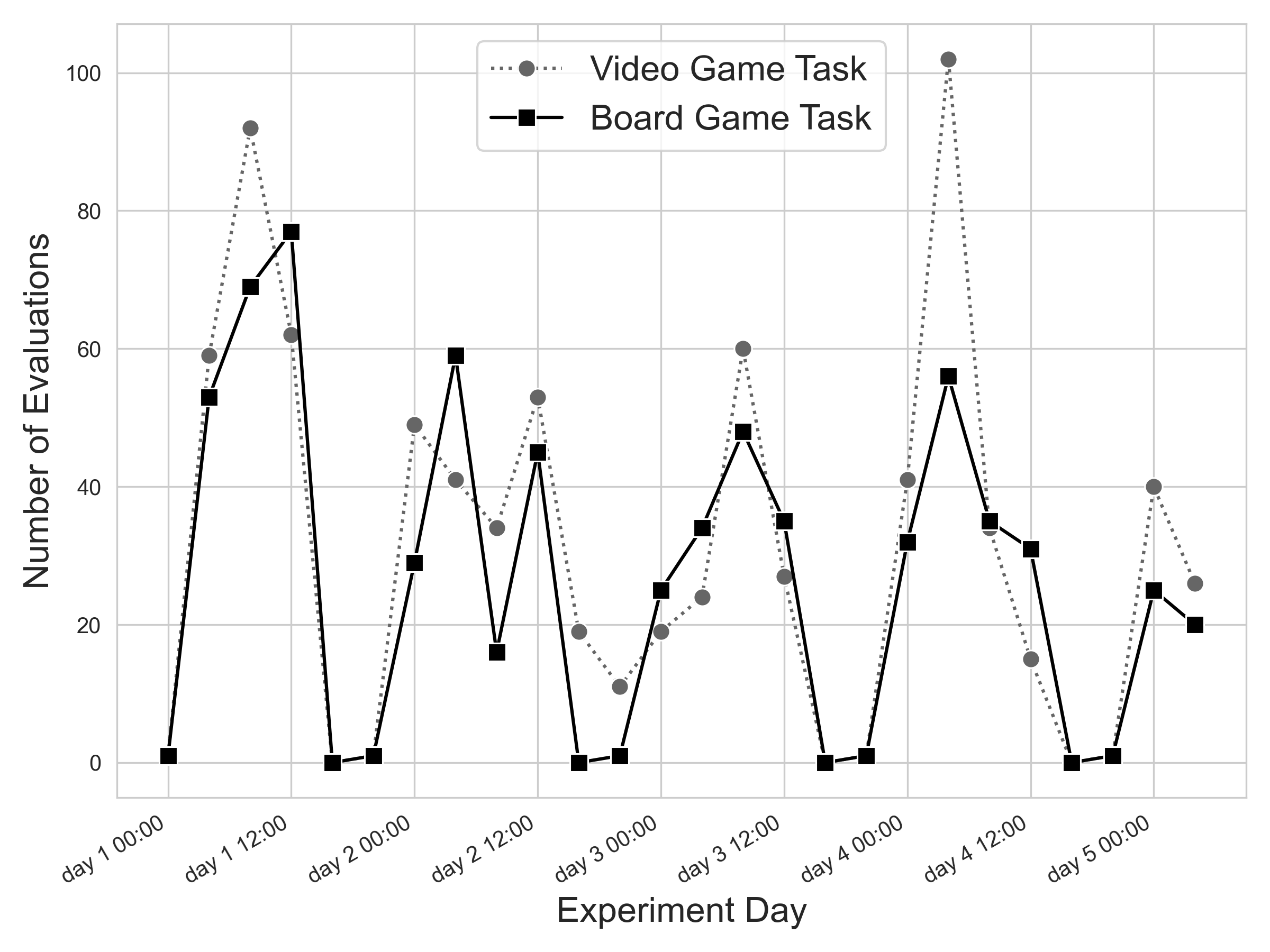}
	\caption{Number of evaluations submitted during the experiments.}
	\label{fig:evaluations}
	
	\centering
	\includegraphics[width=.8\columnwidth]{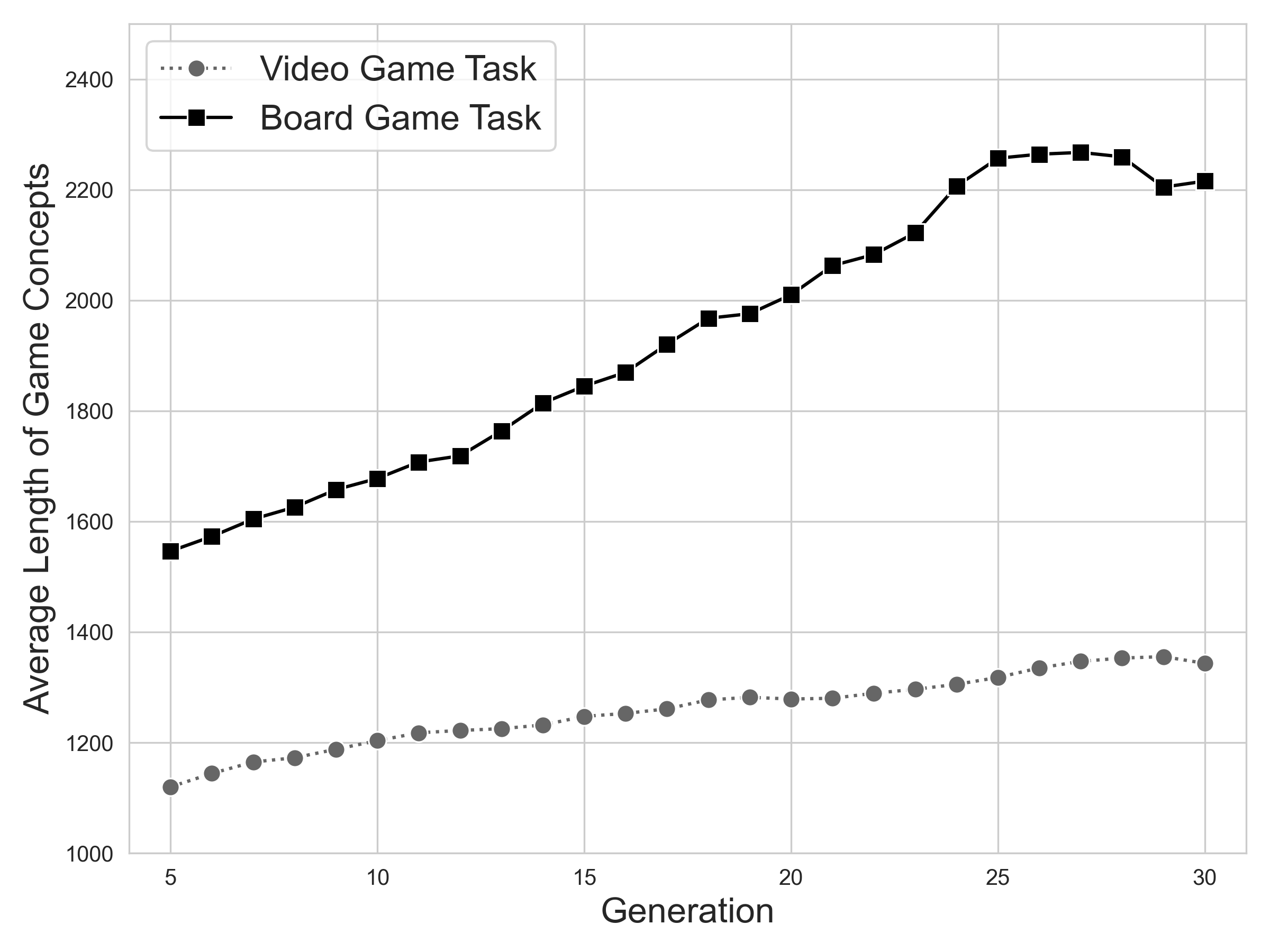}
	\caption{Average length (as number of characters) of the game concepts, as number of characters in the population computed as the moving average over the last five activation of the evolutionary algorithm.}
	\label{fig:concept_length}
	
\end{figure}

Over the four days, the evolutionary algorithm performed 30 iterations and a total of 40 game ideas were evaluated for each task; board game ideas received a total of 799 evaluations; video game ideas received 1025 evaluations overall. Figure~\ref{fig:evaluations} shows the number of evaluations submitted over the four and half days. The plots follow the schedule with peaks around the publication slots and no evaluations during the night time. Figure~\ref{fig:concept_length} shows the average length of the concepts (as the number of characters) in the population computed as the moving average over the last 5 activations of the evolutionary algorithm. As can be noted, board game concepts tend to be longer on average (Figure~\ref{fig:concept_length}) and therefore require more effort to evaluate; as a consequence, the board game concepts received fewer evaluations than the ones of minimalistic video games as Figure~\ref{fig:evaluations} shows.

\begin{figure}
	\centering
	\includegraphics[width=\columnwidth]{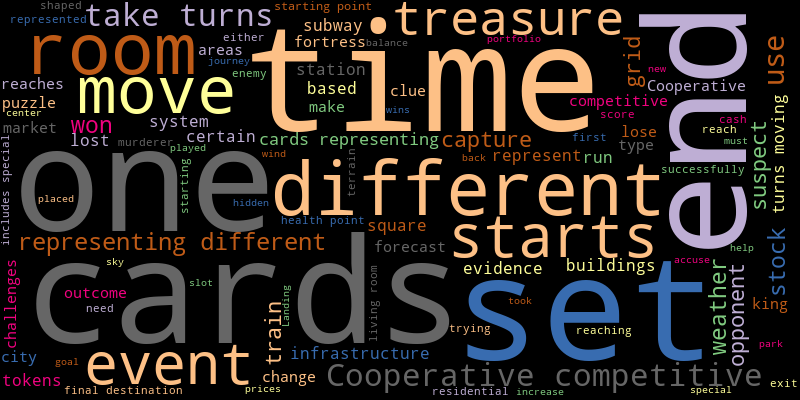}
	
	\centerline{(a)}	
	
	\includegraphics[width=\columnwidth]{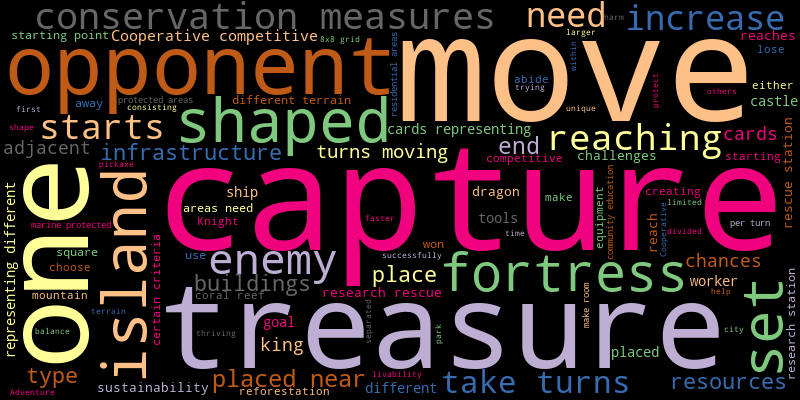}
	
	\centerline{(b)}	
	\caption{Word cloud for (a) the initial population and (b) the final population of the board game design task.}
	\label{fig:boardgame_cloud}	
\end{figure}

Figure~\ref{fig:boardgame_cloud} compares the word cloud generated using all the board game ideas in the original population (\ref{fig:boardgame_cloud}a) with the one generated from the final population (\ref{fig:boardgame_cloud}b). The clouds share some general terms (e.g., moves, starts), but the game concepts in the final population focus on completely different concepts like the terms related to ecosystem maintenance (e.g., island, conservation, resources). 
Due to lack of space, we cannot report the word clouds for the video game task as well as examples of game concepts generated (we refer the reader to~\cite{lanzi2023chatgpt} for additional details).

Finally, we asked participants to complete a form with four open questions. The questions and a summary of the participants' opinions, based on 10 responses collected for the board game task and 13 for the video game task, are presented below.

\smallskip\noindent
\textbf{(i) Which game concepts did they like most, and why?}
Many games in the final population shared similar elements and mechanics.
Although participants did not prefer a single game over the 40 evaluated, there was a common preference for concepts that involved one specific mechanic for each task. 
For the board game design task, most participants liked a cooperative mechanic that asked players to maintain the equilibrium of an ecosystem by placing tiles on a board. 
Interestingly, no elements of this mechanic were present in the initial population but the mechanic emerged later in the experiment in games with different narratives (e.g., one set in an island environment, one in an urban environment). In the video game task, most participants liked a game mechanic involving  \textit{reflection} in its different interpretations (e.g., reflection of light in some games, reflection of bouncing bullets in other ones). In this case, the initial population contained one game description using reflection of bullets however the light reflection mechanics appeared much later in the population. 

\smallskip\noindent
\textbf{(ii) What were the positive features of the generated concepts?}
Participants mainly liked the emergence of novelty in mechanics, player inputs, and unexpected element combinations. For instance, in the video game task, a novel mechanic involving color manipulation combined with shadow and light mechanics positively impressed some participants. Additionally, the emergence of different narratives (e.g., environmental protection, urban planning) and scenarios (e.g., a jungle, an island, etc.) in the board game task was positively received.

\smallskip\noindent
\textbf{(iii) What were the negative features of the generated concepts?}
The two main issues reported were the degree of similarity and incoherence in some generated concepts. The similarity issue resulted from the small population size (10 individuals) and a limited number of steady-state iterations (30), which increased the likelihood of presenting several vations of the same concept. However, this design choice was necessary to reduce the evaluation burden. 
The incoherence issue stemmed from the current limitations of LLMs~\cite{shanahan2022talking,mahowald2023dissociating}.
Despite we might expect future improvements in this respect, the lack of coherence is a common problem in these models and it is not easy to solve.

\smallskip\noindent
\textbf{(iv) How could the collaborative design experience be improved?}
Interestingly, most participants did not suggest increasing their agency but rather improving concept presentation. Some participants recommended presenting concepts more visually (e.g., using images or videos), while others suggested highlighting similarities and differences among the concepts to facilitate evaluation.

\subsection{2023 Global Game Jam}
We performed a final experiment during the 2023 Global Game Jam that ran February 3-5, 2023, from Friday afternoon to Sunday evening. The event always starts at 5 pm on Friday with the publication of the theme (this year was \emph{roots}) and ends Sunday evening with the presentation of the games. The jam is a global event run with a local timing in that it starts at 5 pm and ends Sunday around 6 pm \textit{in every time zone}; jam sites in the Asia continent are the first to start, those in the Hawaiian time zone are the last ones.
We contacted organizers of the Global Game Jam sites in our country, sent them a link to a dedicated Telegram group, and asked them to advertise the tool's availability to the participants of their jam sites; around 35 people joined the group, most of them anonymously. 
In the previous experiments, at the beginning, participants received a brief with a list of constraints and were instructed to cooperate toward a common objective. 
In contrast, in this case, participants had a theme (\emph{roots}) that could lead to several different interpretations; 
they did not have constraints and did not have to collaborate but they brainstormed for ideas to develop their own game.
 In addition, the time window was more limited since the game must be implemented by Sunday afternoon, so teams need to come up with a game idea by Saturday around noon or early afternoon at the latest. When the theme was revealed on February 3 at 5 pm, we published the first ten game ideas randomly generated using the procedure discussed in Section~\ref{sec:evolutionary}; the framework remained online until Saturday night. The evolutionary operators were implemented using DaVinci GPT-3 \cite{gpt3}; the selection was triggered every 5 new evaluations; recombination and mutation probabilities were set as in the previous experiments. The evolutionary algorithm ran for 15 iterations with a number of evaluations much lower than what was recorded in the previous experiments. This is probably due to the short span that jammers can dedicate to the brainstorming process and the in-person nature of the event. Global Game Jam is a physical jam that encourages meeting, working, and sharing ideas with new people. Thus, it is no surprise that an online framework like ours did not fit well in this in-person scenario.  
\section{Conclusions}
We presented an online collaborative design framework combining interactive evolution and large language models for helping users evolve design ideas represented as free-form text. Our framework can support collaboration through traditional web-based grid interfaces \cite{DBLP:conf/chi/SecretanB08,DBLP:conf/gecco/CardamoneLL11} as well as messaging systems, like Telegram, which in our opinion might fit free-text content better. We evaluated our framework in three scenarios involving around 80 participants. Two scenarios mimicked the typical collaborative design process with objectives specified at the start and a fixed duration. The third one was conceived as an open-ended brainstorming activity that took place during the 2023 Global Game Jam. The first two used an evolutionary engine based on ChatGPT; the third one employed DaVinci GPT-3 for the same purpose. All the experiments were run using the Telegram-based interface of our framework. 
Overall, we received positive feedback from the participants. In our experience, we did not notice any difference between the two large language models that are, in our opinion, both viable options. We focused our evaluation on game design since we had access to several junior and senior game designers employed in the industry. However, the framework can be applied to any design task which can be described in free-text form. And if needed, it could be easily extended with text-to-image platforms (like Dall$\cdot$E 2\cite{dall-e-2} or Midjourney\footnote{\url{https://www.midjourney.com}}) to enrich design ideas with captivating visuals. 
\bibliographystyle{ACM-Reference-Format}

\begin{thebibliography}{55}

	%%% the Web address.
	
	\ifx \showCODEN    \undefined \def \showCODEN     #1{\unskip}     \fi
	\ifx \showDOI      \undefined \def \showDOI       #1{#1}\fi
	\ifx \showISBNx    \undefined \def \showISBNx     #1{\unskip}     \fi
	\ifx \showISBNxiii \undefined \def \showISBNxiii  #1{\unskip}     \fi
	\ifx \showISSN     \undefined \def \showISSN      #1{\unskip}     \fi
	\ifx \showLCCN     \undefined \def \showLCCN      #1{\unskip}     \fi
	\ifx \shownote     \undefined \def \shownote      #1{#1}          \fi
	\ifx \showarticletitle \undefined \def \showarticletitle #1{#1}   \fi
	\ifx \showURL      \undefined \def \showURL       {\relax}        \fi
	% invisible to TeX
	\providecommand\bibfield[2]{#2}
	\providecommand\bibinfo[2]{#2}
	\providecommand\natexlab[1]{#1}
	\providecommand\showeprint[2][]{arXiv:#2}
	
	\bibitem[\protect\citeauthoryear{Bontrager, Lin, Togelius, and Risi}{Bontrager
	  et~al\mbox{.}}{2018}]%
			{DBLP:conf/evoW/BontragerLTR18}
	\bibfield{author}{\bibinfo{person}{Philip Bontrager}, \bibinfo{person}{Wending
	  Lin}, \bibinfo{person}{Julian Togelius}, {and} \bibinfo{person}{Sebastian
	  Risi}.} \bibinfo{year}{2018}\natexlab{}.
	\newblock \showarticletitle{Deep Interactive Evolution}. In
	  \bibinfo{booktitle}{\emph{Computational Intelligence in Music, Sound, Art and
	  Design - 7th International Conference, EvoMUSART 2018, Parma, Italy, April
	  4-6, 2018, Proceedings}} \emph{(\bibinfo{series}{Lecture Notes in Computer
	  Science}, Vol.~\bibinfo{volume}{10783})},
	  \bibfield{editor}{\bibinfo{person}{Antonios Liapis}, \bibinfo{person}{Juan
	  Jes{\'{u}}s~Romero Cardalda}, {and} \bibinfo{person}{Anik{\'{o}}
	  Ek{\'{a}}rt}} (Eds.). \bibinfo{publisher}{Springer},
	  \bibinfo{pages}{267--282}.
	\newblock
	\urldef\tempurl%
	\url{https://doi.org/10.1007/978-3-319-77583-8\_18}
	\showDOI{\tempurl}
	
	
	\bibitem[\protect\citeauthoryear{Brintrup, Ramsden, Takagi, and
	  Tiwari}{Brintrup et~al\mbox{.}}{2008}]%
			{brintrup2008ergonomic}
	\bibfield{author}{\bibinfo{person}{A.M. Brintrup}, \bibinfo{person}{J.
	  Ramsden}, \bibinfo{person}{H. Takagi}, {and} \bibinfo{person}{A. Tiwari}.}
	  \bibinfo{year}{2008}\natexlab{}.
	\newblock \showarticletitle{Ergonomic Chair Design by Fusing Qualitative and
	  Quantitative Criteria Using Interactive Genetic Algorithms}.
	\newblock \bibinfo{journal}{\emph{Evolutionary Computation, IEEE Transactions
	  on}} \bibinfo{volume}{12}, \bibinfo{number}{3} (\bibinfo{date}{June}
	  \bibinfo{year}{2008}), \bibinfo{pages}{343 --354}.
	\newblock
	\showISSN{1089-778X}
	\urldef\tempurl%
	\url{https://doi.org/10.1109/TEVC.2007.904343}
	\showDOI{\tempurl}
	
	
	\bibitem[\protect\citeauthoryear{Brown, Horrigan, Eigenfeldt, Gifford, Field,
	  and McCormack}{Brown et~al\mbox{.}}{2018}]%
			{DBLP:conf/nime/BrownHEGFM18}
	\bibfield{author}{\bibinfo{person}{Andrew~R. Brown}, \bibinfo{person}{Matthew
	  Horrigan}, \bibinfo{person}{Arne Eigenfeldt}, \bibinfo{person}{Toby Gifford},
	  \bibinfo{person}{Daniel Field}, {and} \bibinfo{person}{Jon McCormack}.}
	  \bibinfo{year}{2018}\natexlab{}.
	\newblock \showarticletitle{Interacting with Musebots}. In
	  \bibinfo{booktitle}{\emph{18th International Conference on New Interfaces for
	  Musical Expression, {NIME} 2018, Blacksburg, VA, USA, June 3-6, 2018}}.
	  \bibinfo{publisher}{nime.org}, \bibinfo{pages}{19--24}.
	\newblock
	\urldef\tempurl%
	\url{http://www.nime.org/proceedings/2018/nime2018\_paper0004.pdf}
	\showURL{%
	\tempurl}
	
	
	\bibitem[\protect\citeauthoryear{Brown, Mann, Ryder, Subbiah, Kaplan, and
	  al.}{Brown et~al\mbox{.}}{2020}]%
			{gpt3}
	\bibfield{author}{\bibinfo{person}{Tom~B. Brown}, \bibinfo{person}{Benjamin
	  Mann}, \bibinfo{person}{Nick Ryder}, \bibinfo{person}{Melanie Subbiah},
	  \bibinfo{person}{Jared Kaplan}, {and} \bibinfo{person}{al.}}
	  \bibinfo{year}{2020}\natexlab{}.
	\newblock \bibinfo{title}{Language Models are Few-Shot Learners}.
	\newblock
	\newblock
	\urldef\tempurl%
	\url{https://doi.org/10.48550/ARXIV.2005.14165}
	\showDOI{\tempurl}
	
	
	\bibitem[\protect\citeauthoryear{Cardamone, Lanzi, and Loiacono}{Cardamone
	  et~al\mbox{.}}{2015}]%
			{DBLP:journals/asc/CardamoneLL15}
	\bibfield{author}{\bibinfo{person}{Luigi Cardamone}, \bibinfo{person}{Pier~Luca
	  Lanzi}, {and} \bibinfo{person}{Daniele Loiacono}.}
	  \bibinfo{year}{2015}\natexlab{}.
	\newblock \showarticletitle{TrackGen: An interactive track generator for
	  {TORCS} and Speed-Dreams}.
	\newblock \bibinfo{journal}{\emph{Appl. Soft Comput.}}  \bibinfo{volume}{28}
	  (\bibinfo{year}{2015}), \bibinfo{pages}{550--558}.
	\newblock
	\urldef\tempurl%
	\url{https://doi.org/10.1016/j.asoc.2014.11.010}
	\showDOI{\tempurl}
	
	
	\bibitem[\protect\citeauthoryear{Cardamone, Loiacono, and Lanzi}{Cardamone
	  et~al\mbox{.}}{2011}]%
			{DBLP:conf/gecco/CardamoneLL11}
	\bibfield{author}{\bibinfo{person}{Luigi Cardamone}, \bibinfo{person}{Daniele
	  Loiacono}, {and} \bibinfo{person}{Pier~Luca Lanzi}.}
	  \bibinfo{year}{2011}\natexlab{}.
	\newblock \showarticletitle{Interactive evolution for the procedural generation
	  of tracks in a high-end racing game}. In \bibinfo{booktitle}{\emph{13th
	  Annual Genetic and Evolutionary Computation Conference, {GECCO} 2011,
	  Proceedings, Dublin, Ireland, July 12-16, 2011}},
	  \bibfield{editor}{\bibinfo{person}{Natalio Krasnogor} {and}
	  \bibinfo{person}{Pier~Luca Lanzi}} (Eds.). \bibinfo{publisher}{{ACM}},
	  \bibinfo{pages}{395--402}.
	\newblock
	\urldef\tempurl%
	\url{https://doi.org/10.1145/2001576.2001631}
	\showDOI{\tempurl}
	
	
	\bibitem[\protect\citeauthoryear{Chen, Radford, Child, Wu, Jun, Luan, and
	  Sutskever}{Chen et~al\mbox{.}}{2020}]%
			{DBLP:conf/icml/ChenRC0JLS20}
	\bibfield{author}{\bibinfo{person}{Mark Chen}, \bibinfo{person}{Alec Radford},
	  \bibinfo{person}{Rewon Child}, \bibinfo{person}{Jeffrey Wu},
	  \bibinfo{person}{Heewoo Jun}, \bibinfo{person}{David Luan}, {and}
	  \bibinfo{person}{Ilya Sutskever}.} \bibinfo{year}{2020}\natexlab{}.
	\newblock \showarticletitle{Generative Pretraining From Pixels}. In
	  \bibinfo{booktitle}{\emph{Proceedings of the 37th International Conference on
	  Machine Learning, {ICML} 2020, 13-18 July 2020, Virtual Event}}
	  \emph{(\bibinfo{series}{Proceedings of Machine Learning Research},
	  Vol.~\bibinfo{volume}{119})}. \bibinfo{publisher}{{PMLR}},
	  \bibinfo{pages}{1691--1703}.
	\newblock
	\urldef\tempurl%
	\url{http://proceedings.mlr.press/v119/chen20s.html}
	\showURL{%
	\tempurl}
	
	
	\bibitem[\protect\citeauthoryear{Chen, Tworek, Jun, Yuan, Pinto, and al.}{Chen
	  et~al\mbox{.}}{2021}]%
			{https://doi.org/10.48550/arxiv.2107.03374}
	\bibfield{author}{\bibinfo{person}{Mark Chen}, \bibinfo{person}{Jerry Tworek},
	  \bibinfo{person}{Heewoo Jun}, \bibinfo{person}{Qiming Yuan},
	  \bibinfo{person}{Henrique Ponde de~Oliveira Pinto}, {and}
	  \bibinfo{person}{al.}} \bibinfo{year}{2021}\natexlab{}.
	\newblock \bibinfo{title}{Evaluating Large Language Models Trained on Code}.
	\newblock
	\newblock
	\urldef\tempurl%
	\url{https://doi.org/10.48550/ARXIV.2107.03374}
	\showDOI{\tempurl}
	
	
	\bibitem[\protect\citeauthoryear{Cobbe, Kosaraju, Bavarian, Chen, Jun, Kaiser,
	  Plappert, Tworek, Hilton, Nakano, Hesse, and Schulman}{Cobbe
	  et~al\mbox{.}}{2021}]%
			{https://doi.org/10.48550/arxiv.2110.14168}
	\bibfield{author}{\bibinfo{person}{Karl Cobbe}, \bibinfo{person}{Vineet
	  Kosaraju}, \bibinfo{person}{Mohammad Bavarian}, \bibinfo{person}{Mark Chen},
	  \bibinfo{person}{Heewoo Jun}, \bibinfo{person}{Lukasz Kaiser},
	  \bibinfo{person}{Matthias Plappert}, \bibinfo{person}{Jerry Tworek},
	  \bibinfo{person}{Jacob Hilton}, \bibinfo{person}{Reiichiro Nakano},
	  \bibinfo{person}{Christopher Hesse}, {and} \bibinfo{person}{John Schulman}.}
	  \bibinfo{year}{2021}\natexlab{}.
	\newblock \bibinfo{title}{Training Verifiers to Solve Math Word Problems}.
	\newblock
	\newblock
	\urldef\tempurl%
	\url{https://doi.org/10.48550/ARXIV.2110.14168}
	\showDOI{\tempurl}
	
	
	\bibitem[\protect\citeauthoryear{Dawkins}{Dawkins}{1986}]%
			{dawkins:blind_watchmaker}
	\bibfield{author}{\bibinfo{person}{Richard Dawkins}.}
	  \bibinfo{year}{1986}\natexlab{}.
	\newblock \bibinfo{booktitle}{\emph{The Blind Watchmaker: Why the Evidence of
	  Evolution Reveals a Universe without Design}}.
	\newblock \bibinfo{publisher}{Norton \& Company, Inc}.
	\newblock
	\showISBNx{0-393-31570-3}
	
	
	\bibitem[\protect\citeauthoryear{Ebner, Reinhardt, and Albert}{Ebner
	  et~al\mbox{.}}{2005}]%
			{DBLP:conf/eurogp/EbnerRA05}
	\bibfield{author}{\bibinfo{person}{Marc Ebner}, \bibinfo{person}{Markus
	  Reinhardt}, {and} \bibinfo{person}{J{\"{u}}rgen Albert}.}
	  \bibinfo{year}{2005}\natexlab{}.
	\newblock \showarticletitle{Evolution of Vertex and Pixel Shaders}. In
	  \bibinfo{booktitle}{\emph{Genetic Programming, 8th European Conference,
	  EuroGP2005, Lausanne, Switzerland, March 30 - April 1, 2005, Proceedings}}
	  \emph{(\bibinfo{series}{Lecture Notes in Computer Science},
	  Vol.~\bibinfo{volume}{3447})}, \bibfield{editor}{\bibinfo{person}{Maarten
	  Keijzer}, \bibinfo{person}{Andrea Tettamanzi}, \bibinfo{person}{Pierre
	  Collet}, \bibinfo{person}{Jano~I. van Hemert}, {and} \bibinfo{person}{Marco
	  Tomassini}} (Eds.). \bibinfo{publisher}{Springer}, \bibinfo{pages}{261--270}.
	\newblock
	\urldef\tempurl%
	\url{https://doi.org/10.1007/978-3-540-31989-4\_23}
	\showDOI{\tempurl}
	
	
	\bibitem[\protect\citeauthoryear{Frans}{Frans}{2021}]%
			{9619126}
	\bibfield{author}{\bibinfo{person}{Kevin Frans}.}
	  \bibinfo{year}{2021}\natexlab{}.
	\newblock \showarticletitle{AI Charades: Language Models as Interactive Game
	  Environments}. In \bibinfo{booktitle}{\emph{2021 IEEE Conference on Games
	  (CoG)}}. \bibinfo{pages}{1--2}.
	\newblock
	\urldef\tempurl%
	\url{https://doi.org/10.1109/CoG52621.2021.9619126}
	\showDOI{\tempurl}
	
	
	\bibitem[\protect\citeauthoryear{Giacomello, Lanzi, and Loiacono}{Giacomello
	  et~al\mbox{.}}{2019}]%
			{DBLP:conf/cig/GiacomelloLL19}
	\bibfield{author}{\bibinfo{person}{Edoardo Giacomello},
	  \bibinfo{person}{Pier~Luca Lanzi}, {and} \bibinfo{person}{Daniele Loiacono}.}
	  \bibinfo{year}{2019}\natexlab{}.
	\newblock \showarticletitle{Searching the Latent Space of a Generative
	  Adversarial Network to Generate {DOOM} Levels}. In
	  \bibinfo{booktitle}{\emph{{IEEE} Conference on Games, CoG 2019, London,
	  United Kingdom, August 20-23, 2019}}. \bibinfo{publisher}{{IEEE}},
	  \bibinfo{pages}{1--8}.
	\newblock
	\urldef\tempurl%
	\url{https://doi.org/10.1109/CIG.2019.8848011}
	\showDOI{\tempurl}
	
	
	\bibitem[\protect\citeauthoryear{Goodfellow, Pouget{-}Abadie, Mirza, Xu,
	  Warde{-}Farley, Ozair, Courville, and Bengio}{Goodfellow
	  et~al\mbox{.}}{2020}]%
			{DBLP:journals/cacm/GoodfellowPMXWO20}
	\bibfield{author}{\bibinfo{person}{Ian~J. Goodfellow}, \bibinfo{person}{Jean
	  Pouget{-}Abadie}, \bibinfo{person}{Mehdi Mirza}, \bibinfo{person}{Bing Xu},
	  \bibinfo{person}{David Warde{-}Farley}, \bibinfo{person}{Sherjil Ozair},
	  \bibinfo{person}{Aaron~C. Courville}, {and} \bibinfo{person}{Yoshua Bengio}.}
	  \bibinfo{year}{2020}\natexlab{}.
	\newblock \showarticletitle{Generative adversarial networks}.
	\newblock \bibinfo{journal}{\emph{Commun. {ACM}}} \bibinfo{volume}{63},
	  \bibinfo{number}{11} (\bibinfo{year}{2020}), \bibinfo{pages}{139--144}.
	\newblock
	\urldef\tempurl%
	\url{https://doi.org/10.1145/3422622}
	\showDOI{\tempurl}
	
	
	\bibitem[\protect\citeauthoryear{Gray and Brown}{Gray and Brown}{2010}]%
			{gray:2010:gamestorming}
	\bibfield{author}{\bibinfo{person}{Dave Gray} {and} \bibinfo{person}{Sunni
	  Brown}.} \bibinfo{year}{2010}\natexlab{}.
	\newblock \bibinfo{booktitle}{\emph{Gamestorming: A Playbook for Innovators,
	  Rulebreakers, and Changemakers}}.
	\newblock \bibinfo{publisher}{O'Reilly}.
	\newblock
	
	
	\bibitem[\protect\citeauthoryear{Hastings, Guha, , and Stanley}{Hastings
	  et~al\mbox{.}}{2009}]%
			{hastings2010}
	\bibfield{author}{\bibinfo{person}{Erin~J. Hastings}, \bibinfo{person}{Ratan~K.
	  Guha}, \bibinfo{person}{}, {and} \bibinfo{person}{Kenneth~O. Stanley}.}
	  \bibinfo{year}{2009}\natexlab{}.
	\newblock \showarticletitle{Automatic Content Generation in the Galactic Arms
	  Race Video Game}.
	\newblock \bibinfo{journal}{\emph{IEEE Transactions on Computational
	  Intelligence and AI in Games}} \bibinfo{volume}{4}, \bibinfo{number}{1}
	  (\bibinfo{year}{2009}), \bibinfo{pages}{245--263}.
	\newblock
	
	
	\bibitem[\protect\citeauthoryear{Hoover and Stanley}{Hoover and
	  Stanley}{2009}]%
			{DBLP:journals/connection/HooverS09}
	\bibfield{author}{\bibinfo{person}{Amy~K. Hoover} {and}
	  \bibinfo{person}{Kenneth~O. Stanley}.} \bibinfo{year}{2009}\natexlab{}.
	\newblock \showarticletitle{Exploiting functional relationships in musical
	  composition}.
	\newblock \bibinfo{journal}{\emph{Connect. Sci.}} \bibinfo{volume}{21},
	  \bibinfo{number}{2{\&}3} (\bibinfo{year}{2009}), \bibinfo{pages}{227--251}.
	\newblock
	\urldef\tempurl%
	\url{https://doi.org/10.1080/09540090902733871}
	\showDOI{\tempurl}
	
	
	\bibitem[\protect\citeauthoryear{Hoover, Szerlip, and Stanley}{Hoover
	  et~al\mbox{.}}{2011}]%
			{DBLP:conf/gecco/HooverSS11}
	\bibfield{author}{\bibinfo{person}{Amy~K. Hoover}, \bibinfo{person}{Paul~A.
	  Szerlip}, {and} \bibinfo{person}{Kenneth~O. Stanley}.}
	  \bibinfo{year}{2011}\natexlab{}.
	\newblock \showarticletitle{Interactively evolving harmonies through functional
	  scaffolding}. In \bibinfo{booktitle}{\emph{13th Annual Genetic and
	  Evolutionary Computation Conference, {GECCO} 2011, Proceedings, Dublin,
	  Ireland, July 12-16, 2011}}, \bibfield{editor}{\bibinfo{person}{Natalio
	  Krasnogor} {and} \bibinfo{person}{Pier~Luca Lanzi}} (Eds.).
	  \bibinfo{publisher}{{ACM}}, \bibinfo{pages}{387--394}.
	\newblock
	\urldef\tempurl%
	\url{https://doi.org/10.1145/2001576.2001630}
	\showDOI{\tempurl}
	
	
	\bibitem[\protect\citeauthoryear{Hornby and Bongard}{Hornby and
	  Bongard}{2012}]%
			{DBLP:conf/gecco/HornbyB12}
	\bibfield{author}{\bibinfo{person}{Gregory~S. Hornby} {and}
	  \bibinfo{person}{Josh~C. Bongard}.} \bibinfo{year}{2012}\natexlab{}.
	\newblock \showarticletitle{Accelerating human-computer collaborative search
	  through learning comparative and predictive user models}. In
	  \bibinfo{booktitle}{\emph{Genetic and Evolutionary Computation Conference,
	  {GECCO} '12, Philadelphia, PA, USA, July 7-11, 2012}},
	  \bibfield{editor}{\bibinfo{person}{Terence Soule} {and}
	  \bibinfo{person}{Jason~H. Moore}} (Eds.). \bibinfo{publisher}{{ACM}},
	  \bibinfo{pages}{225--232}.
	\newblock
	\urldef\tempurl%
	\url{https://doi.org/10.1145/2330163.2330196}
	\showDOI{\tempurl}
	
	
	\bibitem[\protect\citeauthoryear{Howlett, Colton, and Browne}{Howlett
	  et~al\mbox{.}}{2010}]%
			{pcgfragment}
	\bibfield{author}{\bibinfo{person}{Andrew Howlett}, \bibinfo{person}{Simon
	  Colton}, {and} \bibinfo{person}{Cameron Browne}.}
	  \bibinfo{year}{2010}\natexlab{}.
	\newblock \showarticletitle{Evolving pixel shaders for the prototype video game
	  Subversion}. In \bibinfo{booktitle}{\emph{Proceedings of the 3rd
	  International Symposium on AI and Games - A Symposium at the AISB 2010
	  Convention}} \emph{(\bibinfo{series}{Proceedings of the 3rd International
	  Symposium on AI and Games - A Symposium at the AISB 2010 Convention})}.
	  \bibinfo{pages}{41--46}.
	\newblock
	\showISBNx{1902956907}
	\newblock
	\shownote{AISB Symposium on AI and Games 2010 ; Conference date: 29-03-2010
	  Through 01-04-2010.}
	
	
	\bibitem[\protect\citeauthoryear{Khan, Gunpinar, and Sener}{Khan
	  et~al\mbox{.}}{2019}]%
			{KHAN2019106462}
	\bibfield{author}{\bibinfo{person}{Shahroz Khan}, \bibinfo{person}{Erkan
	  Gunpinar}, {and} \bibinfo{person}{Bekir Sener}.}
	  \bibinfo{year}{2019}\natexlab{}.
	\newblock \showarticletitle{GenYacht: An interactive generative design system
	  for computer-aided yacht hull design}.
	\newblock \bibinfo{journal}{\emph{Ocean Engineering}}  \bibinfo{volume}{191}
	  (\bibinfo{year}{2019}), \bibinfo{pages}{106462}.
	\newblock
	\showISSN{0029-8018}
	\urldef\tempurl%
	\url{https://doi.org/10.1016/j.oceaneng.2019.106462}
	\showDOI{\tempurl}
	
	
	\bibitem[\protect\citeauthoryear{Kim and Cho}{Kim and Cho}{2000}]%
			{kim2000application}
	\bibfield{author}{\bibinfo{person}{Hee-Su Kim} {and} \bibinfo{person}{Sung-Bae
	  Cho}.} \bibinfo{year}{2000}\natexlab{}.
	\newblock \showarticletitle{Application of interactive genetic algorithm to
	  fashion design}.
	\newblock \bibinfo{journal}{\emph{Engineering Applications of Artificial
	  Intelligence}} \bibinfo{volume}{13}, \bibinfo{number}{6}
	  (\bibinfo{year}{2000}), \bibinfo{pages}{635 -- 644}.
	\newblock
	\showISSN{0952-1976}
	\urldef\tempurl%
	\url{https://doi.org/DOI: 10.1016/S0952-1976(00)00045-2}
	\showDOI{\tempurl}
	
	
	\bibitem[\protect\citeauthoryear{Kim and Cho}{Kim and Cho}{2005}]%
			{Kim2005}
	\bibfield{author}{\bibinfo{person}{Hee-Su Kim} {and} \bibinfo{person}{Sung-Bae
	  Cho}.} \bibinfo{year}{2005}\natexlab{}.
	\newblock \bibinfo{booktitle}{\emph{Fashion Design Using Interactive Genetic
	  Algorithm with Knowledge-based Encoding}}.
	\newblock \bibinfo{publisher}{Springer Berlin Heidelberg},
	  \bibinfo{address}{Berlin, Heidelberg}, \bibinfo{pages}{411--434}.
	\newblock
	\showISBNx{978-3-540-44511-1}
	\urldef\tempurl%
	\url{https://doi.org/10.1007/978-3-540-44511-1_19}
	\showDOI{\tempurl}
	
	
	\bibitem[\protect\citeauthoryear{Kowaliw, Dorin, and McCormack}{Kowaliw
	  et~al\mbox{.}}{2012}]%
			{DBLP:journals/tec/KowaliwDM12}
	\bibfield{author}{\bibinfo{person}{Taras Kowaliw}, \bibinfo{person}{Alan
	  Dorin}, {and} \bibinfo{person}{Jon McCormack}.}
	  \bibinfo{year}{2012}\natexlab{}.
	\newblock \showarticletitle{Promoting Creative Design in Interactive
	  Evolutionary Computation}.
	\newblock \bibinfo{journal}{\emph{{IEEE} Trans. Evol. Comput.}}
	  \bibinfo{volume}{16}, \bibinfo{number}{4} (\bibinfo{year}{2012}),
	  \bibinfo{pages}{523--536}.
	\newblock
	\urldef\tempurl%
	\url{https://doi.org/10.1109/TEVC.2011.2166764}
	\showDOI{\tempurl}
	
	
	\bibitem[\protect\citeauthoryear{Kowaliw, McCormack, and Dorin}{Kowaliw
	  et~al\mbox{.}}{2011}]%
			{DBLP:conf/ieeealife/KowaliwMD11}
	\bibfield{author}{\bibinfo{person}{Taras Kowaliw}, \bibinfo{person}{Jon
	  McCormack}, {and} \bibinfo{person}{Alan Dorin}.}
	  \bibinfo{year}{2011}\natexlab{}.
	\newblock \showarticletitle{An interactive electronic art system based on
	  artificial ecosystemics}. In \bibinfo{booktitle}{\emph{2011 {IEEE} Symposium
	  on Artificial Life, {ALIFE} 2011, Paris, France, April 13-15, 2011}}.
	  \bibinfo{publisher}{{IEEE}}, \bibinfo{pages}{162--169}.
	\newblock
	\urldef\tempurl%
	\url{https://doi.org/10.1109/ALIFE.2011.5954645}
	\showDOI{\tempurl}
	
	
	\bibitem[\protect\citeauthoryear{Koza}{Koza}{1992}]%
			{koza:1992}
	\bibfield{author}{\bibinfo{person}{John~R. Koza}.}
	  \bibinfo{year}{1992}\natexlab{}.
	\newblock \bibinfo{booktitle}{\emph{Genetic Programming: On the Programming of
	  Computers by Means of Natural Selection}}.
	\newblock \bibinfo{publisher}{The MIT Press}.
	\newblock
	\showISBNx{9780262527910}
	
	
	\bibitem[\protect\citeauthoryear{Lanzi and Loiacono}{Lanzi and
	  Loiacono}{2023}]%
			{lanzi2023chatgpt}
	\bibfield{author}{\bibinfo{person}{Pier~Luca Lanzi} {and}
	  \bibinfo{person}{Daniele Loiacono}.} \bibinfo{year}{2023}\natexlab{}.
	\newblock \bibinfo{title}{ChatGPT and Other Large Language Models as
	  Evolutionary Engines for Online Interactive Collaborative Game Design}.
	\newblock
	\newblock
	\showeprint[arxiv]{2303.02155}~[cs.AI]
	
	
	\bibitem[\protect\citeauthoryear{LLC and Inc.}{LLC and Inc.}{2023}]%
			{telegram}
	\bibfield{author}{\bibinfo{person}{Telegram~FZ LLC} {and}
	  \bibinfo{person}{Telegram~Messenger Inc.}} \bibinfo{year}{2023}\natexlab{}.
	\newblock \bibinfo{title}{Telegram}.
	\newblock
	\newblock
	\urldef\tempurl%
	\url{https://telegram.org}
	\showURL{%
	\tempurl}
	
	
	\bibitem[\protect\citeauthoryear{Mahowald, Ivanova, Blank, Kanwisher,
	  Tenenbaum, and Fedorenko}{Mahowald et~al\mbox{.}}{2023}]%
			{mahowald2023dissociating}
	\bibfield{author}{\bibinfo{person}{Kyle Mahowald}, \bibinfo{person}{Anna~A.
	  Ivanova}, \bibinfo{person}{Idan~A. Blank}, \bibinfo{person}{Nancy Kanwisher},
	  \bibinfo{person}{Joshua~B. Tenenbaum}, {and} \bibinfo{person}{Evelina
	  Fedorenko}.} \bibinfo{year}{2023}\natexlab{}.
	\newblock \bibinfo{title}{Dissociating language and thought in large language
	  models: a cognitive perspective}.
	\newblock
	\newblock
	\showeprint[arxiv]{2301.06627}~[cs.CL]
	
	
	\bibitem[\protect\citeauthoryear{Mirowski, Mathewson, Pittman, and
	  Evans}{Mirowski et~al\mbox{.}}{2022}]%
			{https://doi.org/10.48550/arxiv.2209.14958}
	\bibfield{author}{\bibinfo{person}{Piotr Mirowski}, \bibinfo{person}{Kory~W.
	  Mathewson}, \bibinfo{person}{Jaylen Pittman}, {and} \bibinfo{person}{Richard
	  Evans}.} \bibinfo{year}{2022}\natexlab{}.
	\newblock \bibinfo{title}{Co-Writing Screenplays and Theatre Scripts with
	  Language Models: An Evaluation by Industry Professionals}.
	\newblock
	\newblock
	\urldef\tempurl%
	\url{https://doi.org/10.48550/ARXIV.2209.14958}
	\showDOI{\tempurl}
	
	
	\bibitem[\protect\citeauthoryear{Mok, Xu, Wang, Fan, Kwok, and Xin}{Mok
	  et~al\mbox{.}}{2013}]%
			{MOK20131442}
	\bibfield{author}{\bibinfo{person}{P.Y. Mok}, \bibinfo{person}{Jie Xu},
	  \bibinfo{person}{X.X. Wang}, \bibinfo{person}{J.T. Fan},
	  \bibinfo{person}{Y.L. Kwok}, {and} \bibinfo{person}{John~H. Xin}.}
	  \bibinfo{year}{2013}\natexlab{}.
	\newblock \showarticletitle{An IGA-based design support system for realistic
	  and practical fashion designs}.
	\newblock \bibinfo{journal}{\emph{Computer-Aided Design}} \bibinfo{volume}{45},
	  \bibinfo{number}{11} (\bibinfo{year}{2013}), \bibinfo{pages}{1442--1458}.
	\newblock
	\showISSN{0010-4485}
	\urldef\tempurl%
	\url{https://doi.org/10.1016/j.cad.2013.06.014}
	\showDOI{\tempurl}
	
	
	\bibitem[\protect\citeauthoryear{Myers}{Myers}{2009}]%
			{DBLP:conf/digra/Myers09}
	\bibfield{author}{\bibinfo{person}{David Myers}.}
	  \bibinfo{year}{2009}\natexlab{}.
	\newblock \showarticletitle{In search of a minimalist game}. In
	  \bibinfo{booktitle}{\emph{Proceedings of the 2009 DiGRA International
	  Conference: Breaking New Ground: Innovation in Games, Play, Practice and
	  Theory, DiGRA 2009, London, UK, September 1-4, 2009}},
	  \bibfield{editor}{\bibinfo{person}{Tanya Krzywinska},
	  \bibinfo{person}{Helen~W. Kennedy}, {and} \bibinfo{person}{Barry Atkins}}
	  (Eds.). \bibinfo{publisher}{Digital Games Research Association}.
	\newblock
	\urldef\tempurl%
	\url{http://www.digra.org/digital-library/publications/in-search-of-a-minimalist-game/}
	\showURL{%
	\tempurl}
	
	
	\bibitem[\protect\citeauthoryear{Nealen, Saltsman, and Boxerman}{Nealen
	  et~al\mbox{.}}{2011}]%
			{DBLP:conf/fdg/NealenSB11}
	\bibfield{author}{\bibinfo{person}{Andrew Nealen}, \bibinfo{person}{Adam
	  Saltsman}, {and} \bibinfo{person}{Eddy Boxerman}.}
	  \bibinfo{year}{2011}\natexlab{}.
	\newblock \showarticletitle{Towards minimalist game design}. In
	  \bibinfo{booktitle}{\emph{Foundations of Digital Games, FDG'11, Bordeaux,
	  France, June 28 - July 1, 2011}}, \bibfield{editor}{\bibinfo{person}{Marc
	  Cavazza}, \bibinfo{person}{Katherine Isbister}, {and}
	  \bibinfo{person}{Charles Rich}} (Eds.). \bibinfo{publisher}{{ACM}},
	  \bibinfo{pages}{38--45}.
	\newblock
	\urldef\tempurl%
	\url{https://doi.org/10.1145/2159365.2159371}
	\showDOI{\tempurl}
	
	
	\bibitem[\protect\citeauthoryear{Norman}{Norman}{2013}]%
			{norman:design}
	\bibfield{author}{\bibinfo{person}{Don Norman}.}
	  \bibinfo{year}{2013}\natexlab{}.
	\newblock \bibinfo{booktitle}{\emph{The Design Of Everyday Things}}.
	\newblock \bibinfo{publisher}{Basic Books}.
	\newblock
	\newblock
	\shownote{Revised edition.}
	
	
	\bibitem[\protect\citeauthoryear{{OpenAI}}{{OpenAI}}{2021}]%
			{ChatGPT_2021}
	\bibfield{author}{\bibinfo{person}{{OpenAI}}.} \bibinfo{year}{2021}\natexlab{}.
	\newblock \bibinfo{title}{{ChatGPT}: a Generative Pre-training Transformer
	  Model for Conversational AI}.
	\newblock
	  \bibinfo{howpublished}{\url{https://openai.com/models/gpt-3-citation/}}.
	\newblock
	
	
	\bibitem[\protect\citeauthoryear{{OpenAI}}{{OpenAI}}{2022}]%
			{dall-e-2}
	\bibfield{author}{\bibinfo{person}{{OpenAI}}.} \bibinfo{year}{2022}\natexlab{}.
	\newblock \bibinfo{title}{DALL·E 2}.
	\newblock \bibinfo{howpublished}{\url{https://openai.com/dall-e-2/}}.
	\newblock
	\newblock
	\shownote{Retrieved 2 February 2023.}
	
	
	\bibitem[\protect\citeauthoryear{Pirovano, Mainetti, and Loiacono}{Pirovano
	  et~al\mbox{.}}{2015}]%
			{DBLP:conf/gamesem/PirovanoML15}
	\bibfield{author}{\bibinfo{person}{Michele Pirovano}, \bibinfo{person}{Renato
	  Mainetti}, {and} \bibinfo{person}{Daniele Loiacono}.}
	  \bibinfo{year}{2015}\natexlab{}.
	\newblock \showarticletitle{Volcano: An interactive sword generator}. In
	  \bibinfo{booktitle}{\emph{2015 {IEEE} Games Entertainment Media Conference,
	  {GEM} 2015, Toronto, ON, Canada, October 14-16, 2015}},
	  \bibfield{editor}{\bibinfo{person}{Elena~G. Bertozzi}, \bibinfo{person}{Bill
	  Kapralos}, \bibinfo{person}{Nahum~D. Gershon}, {and} \bibinfo{person}{Jim~R.
	  Parker}} (Eds.). \bibinfo{publisher}{{IEEE}}, \bibinfo{pages}{1--8}.
	\newblock
	\urldef\tempurl%
	\url{https://doi.org/10.1109/GEM.2015.7377226}
	\showDOI{\tempurl}
	
	
	\bibitem[\protect\citeauthoryear{Polu and Sutskever}{Polu and
	  Sutskever}{2020}]%
			{https://doi.org/10.48550/arxiv.2009.03393}
	\bibfield{author}{\bibinfo{person}{Stanislas Polu} {and} \bibinfo{person}{Ilya
	  Sutskever}.} \bibinfo{year}{2020}\natexlab{}.
	\newblock \bibinfo{title}{Generative Language Modeling for Automated Theorem
	  Proving}.
	\newblock
	\newblock
	\urldef\tempurl%
	\url{https://doi.org/10.48550/ARXIV.2009.03393}
	\showDOI{\tempurl}
	
	
	\bibitem[\protect\citeauthoryear{Radford, Wu, Child, Luan, Amodei, and
	  Sutskever}{Radford et~al\mbox{.}}{2019}]%
			{gpt2}
	\bibfield{author}{\bibinfo{person}{Alec Radford}, \bibinfo{person}{Jeff Wu},
	  \bibinfo{person}{Rewon Child}, \bibinfo{person}{David Luan},
	  \bibinfo{person}{Dario Amodei}, {and} \bibinfo{person}{Ilya Sutskever}.}
	  \bibinfo{year}{2019}\natexlab{}.
	\newblock \showarticletitle{Language Models are Unsupervised Multitask
	  Learners}.
	\newblock  (\bibinfo{year}{2019}).
	\newblock
	\urldef\tempurl%
	\url{https://openai.com/blog/better-language-models/}
	\showURL{%
	\tempurl}
	
	
	\bibitem[\protect\citeauthoryear{Risi, Lehman, D'Ambrosio, Hall, and
	  Stanley}{Risi et~al\mbox{.}}{2012}]%
			{DBLP:conf/aiide/RisiLDHS12}
	\bibfield{author}{\bibinfo{person}{Sebastian Risi}, \bibinfo{person}{Joel
	  Lehman}, \bibinfo{person}{David~B. D'Ambrosio}, \bibinfo{person}{Ryan Hall},
	  {and} \bibinfo{person}{Kenneth~O. Stanley}.} \bibinfo{year}{2012}\natexlab{}.
	\newblock \showarticletitle{Combining Search-Based Procedural Content
	  Generation and Social Gaming in the Petalz Video Game}. In
	  \bibinfo{booktitle}{\emph{AIIDE}}, \bibfield{editor}{\bibinfo{person}{Mark
	  Riedl} {and} \bibinfo{person}{Gita Sukthankar}} (Eds.).
	  \bibinfo{publisher}{The AAAI Press}.
	\newblock
	
	
	\bibitem[\protect\citeauthoryear{Schrum, Gutierrez, Volz, Liu, Lucas, and
	  Risi}{Schrum et~al\mbox{.}}{2020}]%
			{10.1145/3377930.3389821}
	\bibfield{author}{\bibinfo{person}{Jacob Schrum}, \bibinfo{person}{Jake
	  Gutierrez}, \bibinfo{person}{Vanessa Volz}, \bibinfo{person}{Jialin Liu},
	  \bibinfo{person}{Simon Lucas}, {and} \bibinfo{person}{Sebastian Risi}.}
	  \bibinfo{year}{2020}\natexlab{}.
	\newblock \showarticletitle{Interactive Evolution and Exploration within Latent
	  Level-Design Space of Generative Adversarial Networks}. In
	  \bibinfo{booktitle}{\emph{Proceedings of the 2020 Genetic and Evolutionary
	  Computation Conference}} (Canc\'{u}n, Mexico) \emph{(\bibinfo{series}{GECCO
	  '20})}. \bibinfo{publisher}{Association for Computing Machinery},
	  \bibinfo{address}{New York, NY, USA}, \bibinfo{pages}{148–156}.
	\newblock
	\showISBNx{9781450371285}
	\urldef\tempurl%
	\url{https://doi.org/10.1145/3377930.3389821}
	\showDOI{\tempurl}
	
	
	\bibitem[\protect\citeauthoryear{Secretan and Beato}{Secretan and
	  Beato}{2008}]%
			{DBLP:conf/chi/SecretanB08}
	\bibfield{author}{\bibinfo{person}{Jimmy Secretan} {and}
	  \bibinfo{person}{Nicholas Beato}.} \bibinfo{year}{2008}\natexlab{}.
	\newblock \showarticletitle{Picbreeder: evolving pictures collaboratively
	  online}. In \bibinfo{booktitle}{\emph{CHI}}. \bibinfo{pages}{1759--1768}.
	\newblock
	
	
	\bibitem[\protect\citeauthoryear{Shanahan}{Shanahan}{2022}]%
			{shanahan2022talking}
	\bibfield{author}{\bibinfo{person}{Murray Shanahan}.}
	  \bibinfo{year}{2022}\natexlab{}.
	\newblock \bibinfo{title}{Talking About Large Language Models}.
	\newblock
	\newblock
	\showeprint[arxiv]{2212.03551}~[cs.CL]
	
	
	\bibitem[\protect\citeauthoryear{Tabatabaei~Anaraki}{Tabatabaei~Anaraki}{2017}]%
			{10.1007/978-3-319-55750-2_20}
	\bibfield{author}{\bibinfo{person}{Nazanin~Alsadat Tabatabaei~Anaraki}.}
	  \bibinfo{year}{2017}\natexlab{}.
	\newblock \showarticletitle{Fashion Design Aid System with Application of
	  Interactive Genetic Algorithms}. In \bibinfo{booktitle}{\emph{Computational
	  Intelligence in Music, Sound, Art and Design}},
	  \bibfield{editor}{\bibinfo{person}{Jo{\~a}o Correia}, \bibinfo{person}{Vic
	  Ciesielski}, {and} \bibinfo{person}{Antonios Liapis}} (Eds.).
	  \bibinfo{publisher}{Springer International Publishing},
	  \bibinfo{address}{Cham}, \bibinfo{pages}{289--303}.
	\newblock
	\showISBNx{978-3-319-55750-2}
	
	
	\bibitem[\protect\citeauthoryear{Takagi}{Takagi}{2001}]%
			{949485}
	\bibfield{author}{\bibinfo{person}{H. Takagi}.}
	  \bibinfo{year}{2001}\natexlab{}.
	\newblock \showarticletitle{Interactive evolutionary computation: fusion of the
	  capabilities of EC optimization and human evaluation}.
	\newblock \bibinfo{journal}{\emph{Proc. IEEE}} \bibinfo{volume}{89},
	  \bibinfo{number}{9} (\bibinfo{year}{2001}), \bibinfo{pages}{1275--1296}.
	\newblock
	\urldef\tempurl%
	\url{https://doi.org/10.1109/5.949485}
	\showDOI{\tempurl}
	
	
	\bibitem[\protect\citeauthoryear{van Stegeren and Myundefinedliwiec}{van
	  Stegeren and Myundefinedliwiec}{2021}]%
			{10.1145/3472538.3472595}
	\bibfield{author}{\bibinfo{person}{Judith van Stegeren} {and}
	  \bibinfo{person}{Jakub Myundefinedliwiec}.} \bibinfo{year}{2021}\natexlab{}.
	\newblock \showarticletitle{Fine-Tuning GPT-2 on Annotated RPG Quests for NPC
	  Dialogue Generation}. In \bibinfo{booktitle}{\emph{Proceedings of the 16th
	  International Conference on the Foundations of Digital Games}} (Montreal, QC,
	  Canada) \emph{(\bibinfo{series}{FDG '21})}. \bibinfo{publisher}{Association
	  for Computing Machinery}, \bibinfo{address}{New York, NY, USA}, Article
	  \bibinfo{articleno}{2}, \bibinfo{numpages}{8}~pages.
	\newblock
	\showISBNx{9781450384223}
	\urldef\tempurl%
	\url{https://doi.org/10.1145/3472538.3472595}
	\showDOI{\tempurl}
	
	
	\bibitem[\protect\citeauthoryear{Volz, Schrum, Liu, Lucas, Smith, and
	  Risi}{Volz et~al\mbox{.}}{2018}]%
			{DBLP:conf/gecco/VolzSLLSR18}
	\bibfield{author}{\bibinfo{person}{Vanessa Volz}, \bibinfo{person}{Jacob
	  Schrum}, \bibinfo{person}{Jialin Liu}, \bibinfo{person}{Simon~M. Lucas},
	  \bibinfo{person}{Adam~M. Smith}, {and} \bibinfo{person}{Sebastian Risi}.}
	  \bibinfo{year}{2018}\natexlab{}.
	\newblock \showarticletitle{Evolving mario levels in the latent space of a deep
	  convolutional generative adversarial network}. In
	  \bibinfo{booktitle}{\emph{Proceedings of the Genetic and Evolutionary
	  Computation Conference, {GECCO} 2018, Kyoto, Japan, July 15-19, 2018}},
	  \bibfield{editor}{\bibinfo{person}{Hern{\'{a}}n~E. Aguirre} {and}
	  \bibinfo{person}{Keiki Takadama}} (Eds.). \bibinfo{publisher}{{ACM}},
	  \bibinfo{pages}{221--228}.
	\newblock
	\urldef\tempurl%
	\url{https://doi.org/10.1145/3205455.3205517}
	\showDOI{\tempurl}
	
	
	\bibitem[\protect\citeauthoryear{Värtinen, Hämäläinen, and
	  Guckelsberger}{Värtinen et~al\mbox{.}}{2022}]%
			{9980408}
	\bibfield{author}{\bibinfo{person}{Susanna Värtinen}, \bibinfo{person}{Perttu
	  Hämäläinen}, {and} \bibinfo{person}{Christian Guckelsberger}.}
	  \bibinfo{year}{2022}\natexlab{}.
	\newblock \showarticletitle{Generating Role-Playing Game Quests With GPT
	  Language Models}.
	\newblock \bibinfo{journal}{\emph{IEEE Transactions on Games}}
	  (\bibinfo{year}{2022}), \bibinfo{pages}{1--12}.
	\newblock
	\urldef\tempurl%
	\url{https://doi.org/10.1109/TG.2022.3228480}
	\showDOI{\tempurl}
	
	
	\bibitem[\protect\citeauthoryear{Wang and Zhou}{Wang and Zhou}{2020}]%
			{WANG2020102901}
	\bibfield{author}{\bibinfo{person}{Tianxiong Wang} {and} \bibinfo{person}{Meiyu
	  Zhou}.} \bibinfo{year}{2020}\natexlab{}.
	\newblock \showarticletitle{A method for product form design of integrating
	  interactive genetic algorithm with the interval hesitation time and user
	  satisfaction}.
	\newblock \bibinfo{journal}{\emph{International Journal of Industrial
	  Ergonomics}}  \bibinfo{volume}{76} (\bibinfo{year}{2020}),
	  \bibinfo{pages}{102901}.
	\newblock
	\showISSN{0169-8141}
	\urldef\tempurl%
	\url{https://doi.org/10.1016/j.ergon.2019.102901}
	\showDOI{\tempurl}
	
	
	\bibitem[\protect\citeauthoryear{Wikipedia}{Wikipedia}{2022}]%
			{wiki:design_brief}
	\bibfield{author}{\bibinfo{person}{Wikipedia}.}
	  \bibinfo{year}{2022}\natexlab{}.
	\newblock \bibinfo{title}{{D}esign {B}rief ({W}ikipedia{,} The Free
	  Encyclopedia)}.
	\newblock
	\newblock
	\urldef\tempurl%
	\url{https://en.wikipedia.org/wiki/Design_brief}
	\showURL{%
	\tempurl}
	
	
	\bibitem[\protect\citeauthoryear{Woolley and Stanley}{Woolley and
	  Stanley}{2014}]%
			{10.1145/2576768.2598353}
	\bibfield{author}{\bibinfo{person}{Brian~G. Woolley} {and}
	  \bibinfo{person}{Kenneth~O. Stanley}.} \bibinfo{year}{2014}\natexlab{}.
	\newblock \showarticletitle{A Novel Human-Computer Collaboration: Combining
	  Novelty Search with Interactive Evolution}. In
	  \bibinfo{booktitle}{\emph{Proceedings of the 2014 Annual Conference on
	  Genetic and Evolutionary Computation}} (Vancouver, BC, Canada)
	  \emph{(\bibinfo{series}{GECCO '14})}. \bibinfo{publisher}{Association for
	  Computing Machinery}, \bibinfo{address}{New York, NY, USA},
	  \bibinfo{pages}{233–240}.
	\newblock
	\showISBNx{9781450326629}
	\urldef\tempurl%
	\url{https://doi.org/10.1145/2576768.2598353}
	\showDOI{\tempurl}
	
	
	\bibitem[\protect\citeauthoryear{Xu, Wang, and Li}{Xu et~al\mbox{.}}{2010}]%
			{xu2010emotional}
	\bibfield{author}{\bibinfo{person}{Bin Xu}, \bibinfo{person}{Shangfei Wang},
	  {and} \bibinfo{person}{Xian Li}.} \bibinfo{year}{2010}\natexlab{}.
	\newblock \showarticletitle{An emotional harmony generation system}. In
	  \bibinfo{booktitle}{\emph{Evolutionary Computation (CEC), 2010 IEEE Congress
	  on}}. \bibinfo{pages}{1 --7}.
	\newblock
	\urldef\tempurl%
	\url{https://doi.org/10.1109/CEC.2010.5586210}
	\showDOI{\tempurl}
	
	
	\bibitem[\protect\citeauthoryear{Yannakakis and Togelius}{Yannakakis and
	  Togelius}{2011}]%
			{5740836}
	\bibfield{author}{\bibinfo{person}{Georgios~N. Yannakakis} {and}
	  \bibinfo{person}{Julian Togelius}.} \bibinfo{year}{2011}\natexlab{}.
	\newblock \showarticletitle{Experience-Driven Procedural Content Generation}.
	\newblock \bibinfo{journal}{\emph{IEEE Transactions on Affective Computing}}
	  \bibinfo{volume}{2}, \bibinfo{number}{3} (\bibinfo{year}{2011}),
	  \bibinfo{pages}{147--161}.
	\newblock
	\urldef\tempurl%
	\url{https://doi.org/10.1109/T-AFFC.2011.6}
	\showDOI{\tempurl}
	
	
	\bibitem[\protect\citeauthoryear{Zhu and Luo}{Zhu and Luo}{2022}]%
			{https://doi.org/10.48550/arxiv.2204.09658}
	\bibfield{author}{\bibinfo{person}{Qihao Zhu} {and} \bibinfo{person}{Jianxi
	  Luo}.} \bibinfo{year}{2022}\natexlab{}.
	\newblock \bibinfo{title}{Generative Design Ideation: A Natural Language
	  Generation Approach}.
	\newblock
	\newblock
	\urldef\tempurl%
	\url{https://doi.org/10.48550/ARXIV.2204.09658}
	\showDOI{\tempurl}
	
	
	\bibitem[\protect\citeauthoryear{Ølsted, Ma, and Risi}{Ølsted
	  et~al\mbox{.}}{2015}]%
			{7257069}
	\bibfield{author}{\bibinfo{person}{Peter~Thorup Ølsted},
	  \bibinfo{person}{Benjamin Ma}, {and} \bibinfo{person}{Sebastian Risi}.}
	  \bibinfo{year}{2015}\natexlab{}.
	\newblock \showarticletitle{Interactive evolution of levels for a competitive
	  multiplayer FPS}. In \bibinfo{booktitle}{\emph{2015 IEEE Congress on
	  Evolutionary Computation (CEC)}}. \bibinfo{pages}{1527--1534}.
	\newblock
	\urldef\tempurl%
	\url{https://doi.org/10.1109/CEC.2015.7257069}
	\showDOI{\tempurl}
	
	
	\end{thebibliography}

\appendix

\clearpage
\section{Examples of Evolved Game Concepts}
\begin{table*}[!h]
	\begin{footnotesize}
		\begin{tabular}{c}\hline
			\\
			\parbox{.8\textwidth}{

			\textbf{Name of the game}: Eco-Island\\
			
			\noindent \textbf{Number of players}: 2-4. Cooperative/competitive game: Cooperative.\\
			
			\noindent \textbf{Game Board}: a square board with 64 tiles in an 8x8 grid, representing a tropical island. Each tile has a different terrain, such as a beach, a rainforest, a volcano, a coral reef and a mangrove. The board also includes a starting point and a research station tile.\\
			
			\noindent \textbf{Game Pieces}: Each player starts with a pawn-shaped game piece, representing their scientist and a set of worker pieces, which are small circles. The players also have a set of cards representing different research equipment, such as a microscope, a GPS, a sample container and a camera, and different conservation measures, such as marine protected areas, reforestation and community education.\\
			
			\noindent \textbf{Rules}:\\
			\noindent
Players take turns moving their game pieces on the board, starting from the starting point and trying to reach the research station tile. Each tile has its own set of environmental issues, such as habitat destruction, pollution, and overfishing. Landing on a tile can either help or harm the player’s research. Players can use their cards to study and mitigate the issues and increase their chances of reaching the research station. Players can also place their conservation measures on any tile, but they have to abide by certain rules such as: Marine protected areas need to be placed near coral reefs to protect them; Reforestation needs to be placed in areas with high soil erosion; Community education needs to be placed near vulnerable communities; Players also have a limited number of workers, they can place on the board to help them implement conservation measures faster. Players can also form alliances with other players to increase their chances of reaching the research station, but they can also betray their allies to gain an advantage. Players can also choose to remove their conservation measures to make room for new measures. The game ends when one player reaches the research station tile or all the tiles on the board are filled with conservation measures.\\
			
			\noindent \textbf{Objective}: The game is won by the player who reaches the research station tile first and has implemented conservation measures that meet certain criteria for biodiversity and sustainability. The players lose if the island becomes unsustainable or all players run out of cards before reaching the research station tile.			

			} \\
			\\
			\hline
			\\
			(a) \\		
			\\\hline
			\\
			\parbox{.8\textwidth}{

			\textbf{Name of the game}: Fortress Siege\\
			
			\noindent \textbf{Number of players}: 2-4. Cooperative/competitive game: Competitive.\\
			
			\noindent \textbf{Game Board}: a rectangular board with 36 tiles in a 6x6 grid. The board is divided into two sections, one for each player's fortress. The fortresses are separated by a neutral zone of 4 tiles.\\
			
			\noindent \textbf{Game Pieces}:  Each player starts with 8 game pieces, consisting of: 1 King piece (larger than others and unique in shape),2 Knight pieces (shaped like horses), 2 Archer pieces (shaped like bows), and 3 Peasant pieces (shaped like small houses).\\
			
			\noindent \textbf{Rules}:\\
			\noindent
			Players take turns moving their pieces one tile at a time within their own fortress. They can move their pieces to any tile within their fortress, except for the king piece which can only move to 1 tile per turn. The goal of the game is to move the king piece to the opponent's fortress while protecting it from enemy pieces. The Knight pieces can move and capture any enemy piece in the adjacent tiles. The Archer pieces can capture any enemy piece in a straight line up to 2 tiles away, and Peasants can capture any enemy piece in an adjacent tile. Each turn, players can either move one of their pieces or capture an opponent's piece with one of their pieces. A player can move any of their pieces, but can only capture an opponent's piece with a piece of the same type. The player who captures the opponent's king piece wins the game.\\
			
			\noindent \textbf{Objective}: Objective: The game ends when one player successfully captures the opponent's king piece. That player wins, and the other players lose.		
			}\\
			\\\hline
			\\
			(b)\\
			\\\hline
			\\
\end{tabular}
	\end{footnotesize}
	
	\caption{Example of evolved games for the board game design task.}
	\label{tab:game_examples_bg}
\end{table*}

\begin{table*}[!h]
	\begin{footnotesize}
		\begin{tabular}{c}\hline
			\\
			\parbox{.8\textwidth}{

			\textbf{Name of the game}: Reflectron\\
			 
			\noindent \textbf{Game Concept}: The game is a minimalist puzzle platformer where the player controls a character that can reflect both light beams and projectiles to interact with the environment and solve puzzles. The goal of the game is to reach the end of each level by reflecting both light beams and projectiles to solve puzzles and overcome obstacles. The core mechanics involve using the arrow keys to move the character, the left mouse button to reflect incoming projectiles, and the right mouse button to manipulate the angle of reflection of incoming light beams.\\
			
			\noindent \textbf{Game resources}: The game only requires a simple background, a character that can reflect incoming projectiles and light beams, objects that can be interacted with by reflecting both projectiles and light beams, and obstacles that can be overcome by reflecting both projectiles and light beams. Player input is required through the use of the arrow keys, left mouse button, and right mouse button.\\
			
			\noindent \textbf{Level design}: The levels are designed by creating obstacles and challenges that require the player to reflect both projectiles and light beams in different ways to overcome them, adding diversity to the gameplay.\\
			
			\noindent \textbf{Game instructions}: Use the arrow keys to move the character, left mouse button to reflect incoming projectiles, and the right mouse button to manipulate the angle of reflection of incoming light beams. Reach the end of each level by reflecting both projectiles and light beams to solve puzzles and overcome obstacles.

			} \\
			\\
			\hline
			\\
			(a) \\		
			\\\hline
			\\
			\parbox{.8\textwidth}{

			\textbf{Name of the game}: SoundEchoes: Kinetic\\
			 
			\noindent \textbf{Game Concept}: The game is a minimalist puzzle game where the player controls a character that can manipulate sound and create echoes to interact with the environment and solve puzzles. The goal of the game is to reach the end of each level by manipulating sound and creating echoes to solve puzzles and overcome obstacles. The core mechanics involve using hand gestures (captured by a camera) to move the character, creating echoes of themselves and manipulating sound, such as increasing or decreasing volume, changing pitch and frequency, and amplitude of sound waves.\\
			
			\noindent \textbf{Game resources}: The game only requires a simple background, a character that can manipulate sound and create echoes, objects that can be interacted with by manipulating sound and echoes, and a camera for capturing hand gestures.\\
			
			\noindent \textbf{Level design}: The levels are designed by creating obstacles and puzzles that require the player to manipulate sound and create echoes in different ways to solve them, adding diversity to the gameplay.\\
			
			\noindent \textbf{Game instructions}: Use hand gestures (captured by a camera) to move the character, create echoes of themselves, and manipulate sound, such as increasing or decreasing volume, changing pitch and frequency, and amplitude of sound waves. Reach the end of each level by manipulating sound and creating echoes to solve puzzles and overcome obstacles.
				
			}\\
			\\\hline
			\\
			(b)\\
			\\\hline
			\\
\end{tabular}
	\end{footnotesize}
	
	\caption{Example of evolved games for the video game design task.}
	\label{tab:game_examples_vg}
\end{table*} 

\begin{table*}[!h]
	\begin{footnotesize}
		\begin{tabular}{c}\hline
			\\
			\parbox{.9\textwidth}{
				A white pixel is the minimum amount of information we can show on-screen, and pressing a key (or a button) is the least interaction we can ask players.Given these two examples of minimalist game.\\

\noindent
\textbf{EXAMPLE1}\\

\noindent
\textbf{Name of the game}: Soundscape

\noindent
\textbf{Game Concept}: The game is a minimalist puzzle game where the player controls a character that can manipulate sound to interact with the environment and solve puzzles. The goal of the game is to reach the end of each level by manipulating sound to solve puzzles and overcome obstacles. The core mechanics involve using the arrow keys to move the character and the spacebar to manipulate sound, such as increasing or decreasing volume, changing pitch and frequency.

\noindent
\textbf{Game resources}: The game only requires a simple background, a character that can manipulate sound and objects that can be interacted with by manipulating sound. Player input is required through the use of the arrow keys and spacebar.

\noindent
\textbf{Level design}: The levels are designed by creating obstacles and puzzles that require the player to manipulate sound in different ways to solve them, adding diversity to the gameplay.

\noindent
\textbf{Game instructions}: Use the arrow keys to move the character and the spacebar to manipulate sound, such as increasing or decreasing volume, changing pitch and frequency. Reach the end of each level by manipulating sound to solve puzzles and overcome obstacles.\\

\noindent
\textbf{EXAMPLE2}\\

\noindent
\textbf{Name of the game}: "Echoes"

\noindent
\textbf{Game Concept}: The game is a minimalist puzzle-platformer where the player controls a character that can create echoes of themselves to interact with the environment and solve puzzles. The goal of the game is to reach the end of each level by creating echoes of themselves to solve puzzles and overcome obstacles. The core mechanics involve using the arrow keys to move the character and the spacebar to create echoes of themselves.

\noindent
\textbf{Game resources}: The game only requires a simple background, a character that can create echoes, and objects that can be interacted with by echoes. Player input is required through the use of the arrow keys and spacebar.

\noindent
\textbf{Level design}: The levels are designed by creating obstacles and challenges that require the player to create echoes in different ways to overcome them, adding diversity to the gameplay.

\noindent
\textbf{Game instructions}: Use the arrow keys to move the character and the spacebar to create echoes of themselves. Reach the end of each level by creating echoes of themselves to solve puzzles and overcome obstacles.\\

\noindent
Act as a game designer and recombine these two games to create a novel minimalist game. In your response do not include any introduction or final comment. In the response, please avoid references to the two games recombined.
			}\\
			\hline
			\\
			(a) \\		
			\\\hline
			\\
			\parbox{.9\textwidth}{
			\textbf{Name of the game}: "EchoSound"\\

			\noindent
			\textbf{Game Concept}: The game is a minimalist puzzle-platformer where the player controls a character that can create echoes of themselves and manipulate sound to interact with the environment and solve puzzles. The goal of the game is to reach the end of each level by creating echoes of themselves and manipulating sound to solve puzzles and overcome obstacles. The core mechanics involve using the arrow keys to move the character, the spacebar to create echoes of themselves, and the shift key to manipulate sound, such as increasing or decreasing volume, changing pitch and frequency.\\

			\noindent
			\textbf{Game resources}: The game only requires a simple background, a character that can create echoes and manipulate sound, and objects that can be interacted with by echoes and sound manipulation. Player input is required through the use of the arrow keys, spacebar, and shift key.\\

			\noindent
			\textbf{Level design}: The levels are designed by creating obstacles and challenges that require the player to create echoes and manipulate sound in different ways to overcome them, adding diversity to the gameplay.\\

			\noindent
			\textbf{Game instructions}: Use the arrow keys to move the character, the spacebar to create echoes of themselves, and the shift key to manipulate sound, such as increasing or decreasing volume, changing pitch and frequency. Reach the end of each level by creating echoes of themselves and manipulating sound to solve puzzles and overcome obstacles.
			}\\
			\\\hline
			\\
			(b)\\
			
	\end{tabular}
	\end{footnotesize}
	\caption{An example of the recombination operator: (a) prompt and (b) response.}
	\label{tab:crossover_example}
\end{table*}

\begin{table*}[!h]
	\begin{footnotesize}
		\begin{tabular}{c}\hline
			\\
			\parbox{.9\textwidth}{
				A white pixel is the minimum amount of information we can show on-screen, and pressing a key (or a button) is the least interaction we can ask players.
				Given the following example of minimalist game.\\
				
				\noindent
				\textbf{EXAMPLE}\\

				\noindent
				\textbf{Name of the game}: "EchoScape"
				
				\noindent
				\textbf{Game Concept}: The game is a minimalist puzzle-platformer where the player controls a character that can manipulate sound and create echoes to interact with the environment and solve puzzles. The goal of the game is to reach the end of each level by manipulating sound and creating echoes to solve puzzles and overcome obstacles. The core mechanics involve using the arrow keys to move the character and the spacebar to manipulate sound, such as increasing or decreasing volume, changing pitch and frequency and create echoes of the character.
				
				\noindent
				\textbf{Game resources}: The game only requires a simple background, a character that can manipulate sound and create echoes, and objects that can be interacted with by manipulating sound and echoes. Player input is required through the use of the arrow keys and spacebar.
				
				\noindent
				\textbf{Level design}: The levels are designed by creating obstacles and challenges that require the player to manipulate sound and create echoes in different ways to overcome them, adding diversity to the gameplay.
				
				\noindent
				\textbf{Game instructions}: Use the arrow keys to move the character and the spacebar to manipulate sound and create echoes of themselves. Reach the end of each level by manipulating sound and creating echoes to solve puzzles and overcome obstacles.\\
				
				Act as a game designer and design a novel minimalist game as a variation of the given example, by changing only the level design. In your response do not include any introduction or final comment. In the response, please avoid explicit reference to the original game.
			}\\
			\hline
			\\
			(a) \\		
			\\\hline
			\\
			\parbox{.9\textwidth}{
			\textbf{Name of the game}: "EchoScape: Echoes in Time"\\
			
			\noindent
			\textbf{Game Concept}: The game is a minimalist puzzle-platformer where the player controls a character that can manipulate sound and create echoes to interact with the environment and solve puzzles. The goal of the game is to reach the end of each level by manipulating sound and creating echoes to solve puzzles and overcome obstacles. The core mechanics involve using the arrow keys to move the character and the spacebar to manipulate sound, such as increasing or decreasing volume, changing pitch and frequency and create echoes of the character.\\
			
			\noindent
			\textbf{Game resources}: The game only requires a simple background, a character that can manipulate sound and create echoes, and objects that can be interacted with by manipulating sound and echoes. Player input is required through the use of the arrow keys and spacebar.\\
			
			\noindent
			\textbf{Level design}: The levels are designed as temporal puzzles where player must manipulate sound and create echoes to interact with the environment and solve puzzles in different eras, adding diversity to the gameplay.\\
			
			\noindent
			\textbf{Game instructions}: Use the arrow keys to move the character and the spacebar to manipulate sound and create echoes of themselves. Reach the end of each level by manipulating sound and creating echoes to solve temporal puzzles and overcome obstacles.\\
			}\\
			\\\hline
			\\
			(b)\\			
	\end{tabular}
	\end{footnotesize}
	\caption{An example of the mutation operator: (a) prompt and (b) response.}
	\label{tab:mutation_example}
\end{table*}

\clearpage
\section{Wordclouds}
\begin{figure*}[!h]
	\begin{tabular}{cc}
		\parbox{.45\textwidth}{
			\includegraphics[width=.45\textwidth]{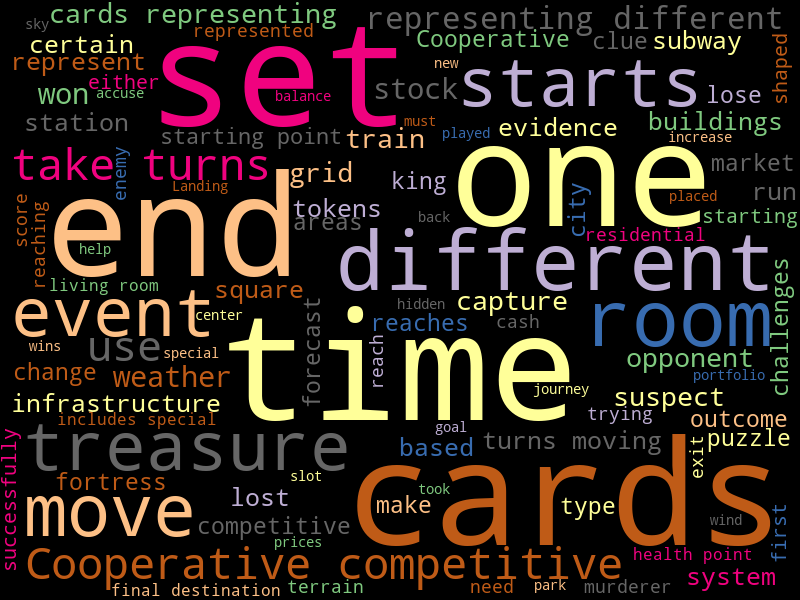}
		}
		
		& 
		\parbox{.45\textwidth}{
			\includegraphics[width=.45\textwidth]{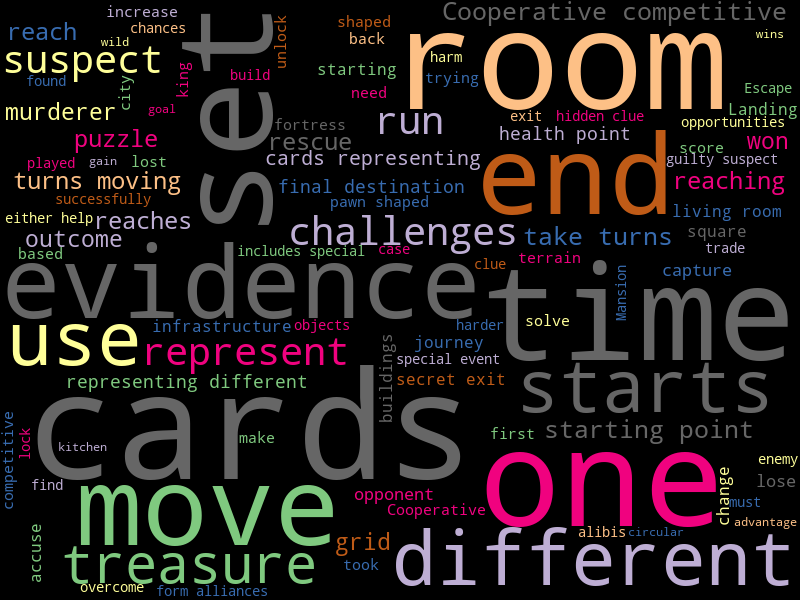}
		} \\
		
		(a) & (b) \\
		
		\parbox{.45\textwidth}{
			\includegraphics[width=.45\textwidth]{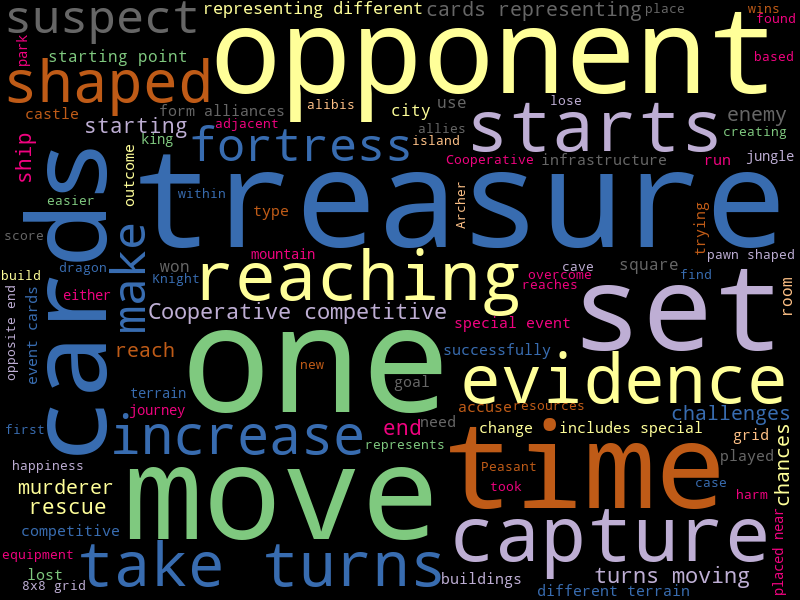}
		}
		
		& 
		\parbox{.45\textwidth}{
			\includegraphics[width=.45\textwidth]{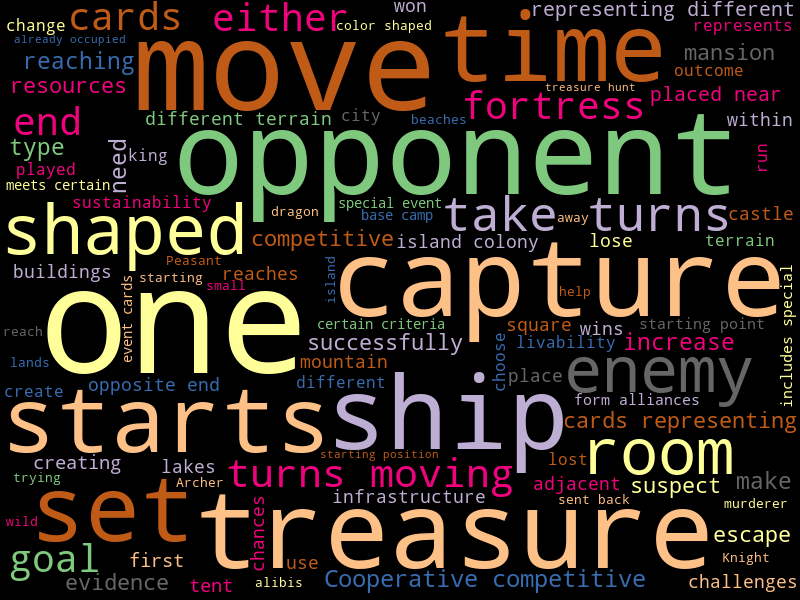}
		} \\
		
		(c) & (d) \\

		\parbox{.45\textwidth}{
			\includegraphics[width=.45\textwidth]{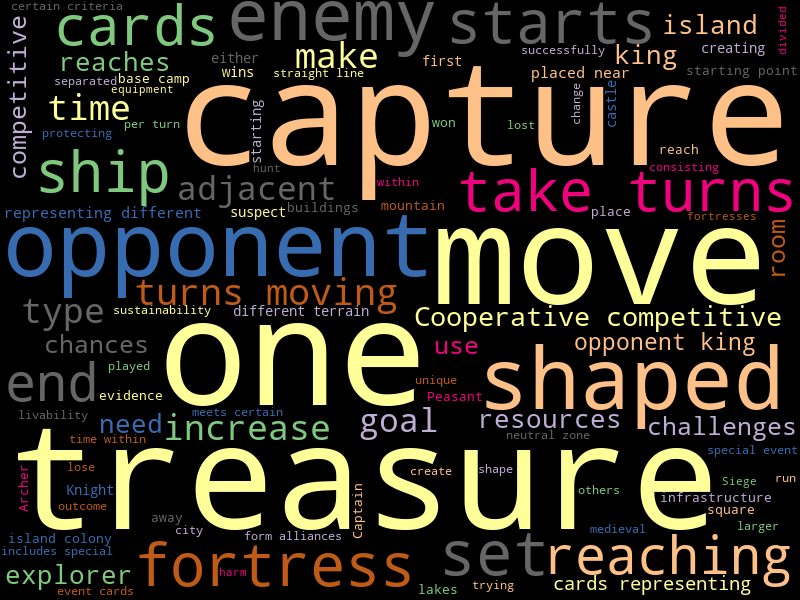}
		}
		
		& 
		\parbox{.45\textwidth}{
			\includegraphics[width=.45\textwidth]{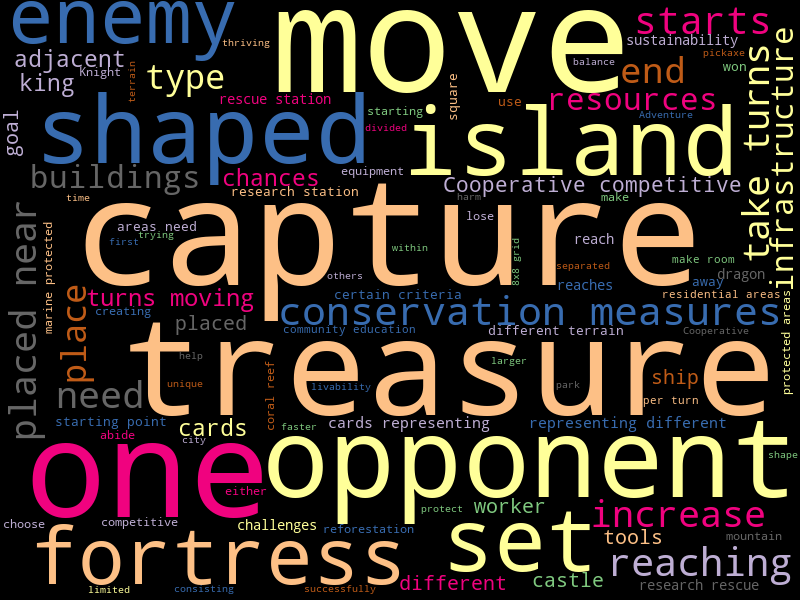}
		} \\
		
		(e) & (f) \\
	\end{tabular}
	\caption{Sequence of word clouds for the board game design task starting from 
		(a) the initial population to the (f) the final population.}
	\label{fig:boardgame_cloud2}	
\end{figure*}

\clearpage
\begin{figure*}[!h]
	\begin{tabular}{cc}
		\parbox{.45\textwidth}{
			\includegraphics[width=.45\textwidth]{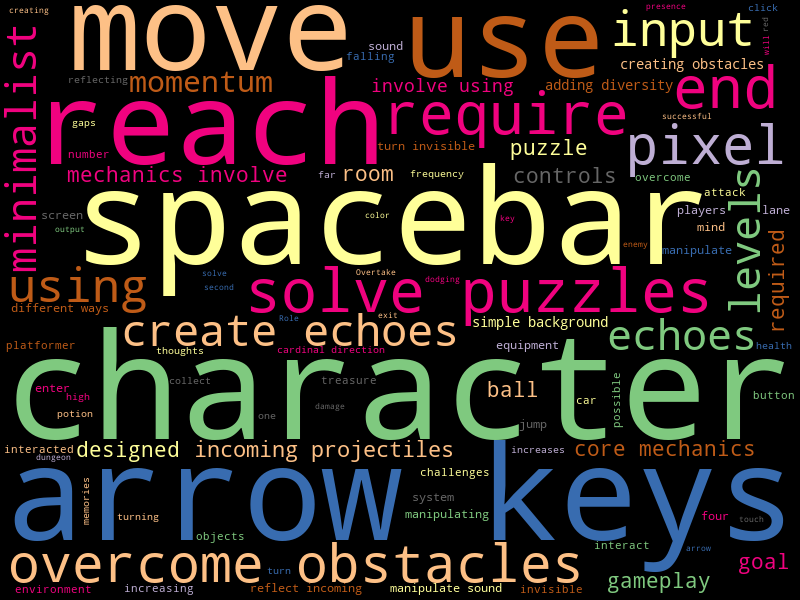}
		}
		
		& 
		\parbox{.45\textwidth}{
			\includegraphics[width=.45\textwidth]{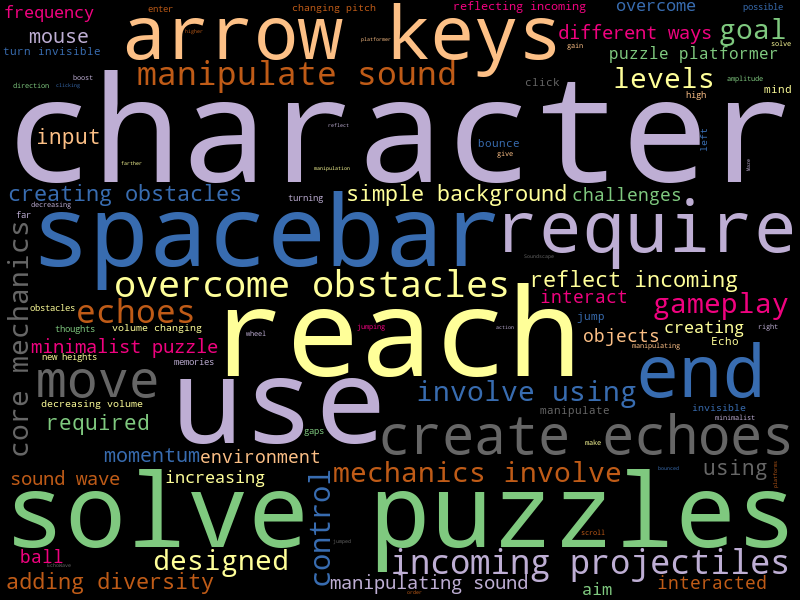}
		} \\
		
		(a) & (b) \\
		
		\parbox{.45\textwidth}{
			\includegraphics[width=.45\textwidth]{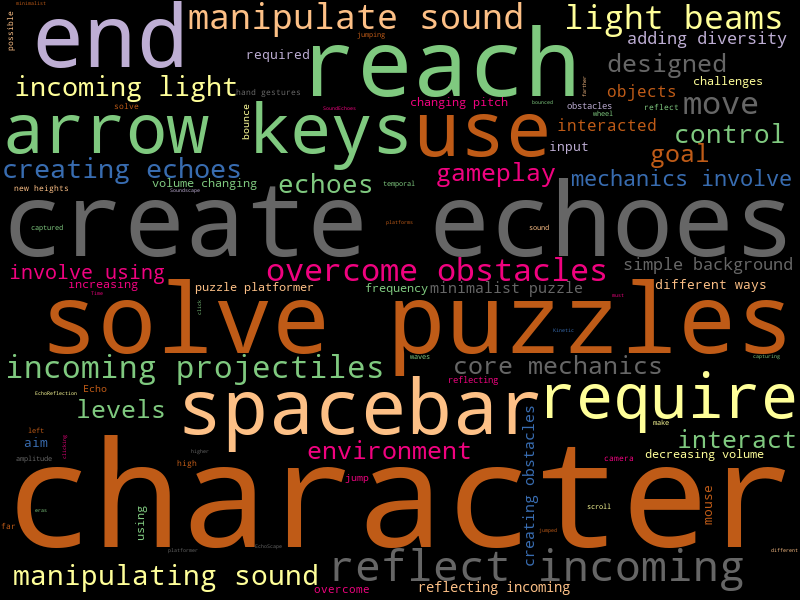}
		}
		
		& 
		\parbox{.45\textwidth}{
			\includegraphics[width=.45\textwidth]{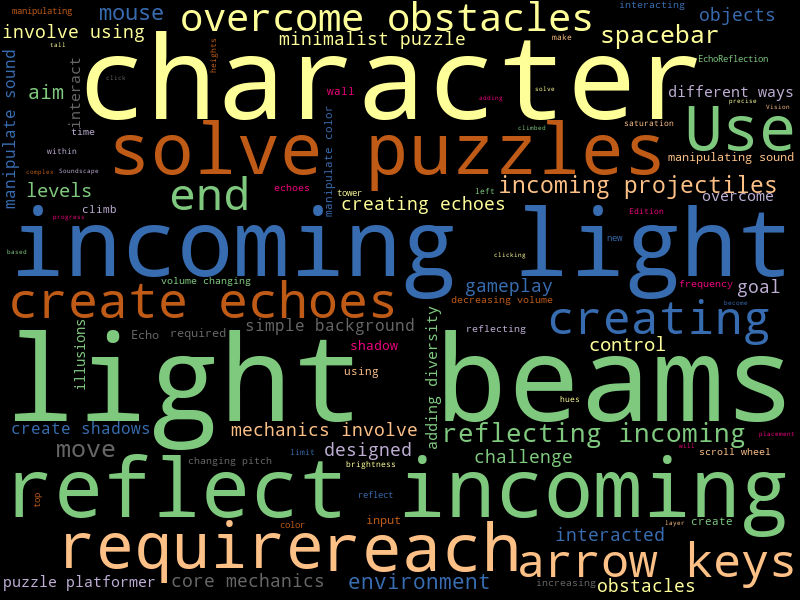}
		} \\
		
		(c) & (d) \\

		\parbox{.45\textwidth}{
			\includegraphics[width=.45\textwidth]{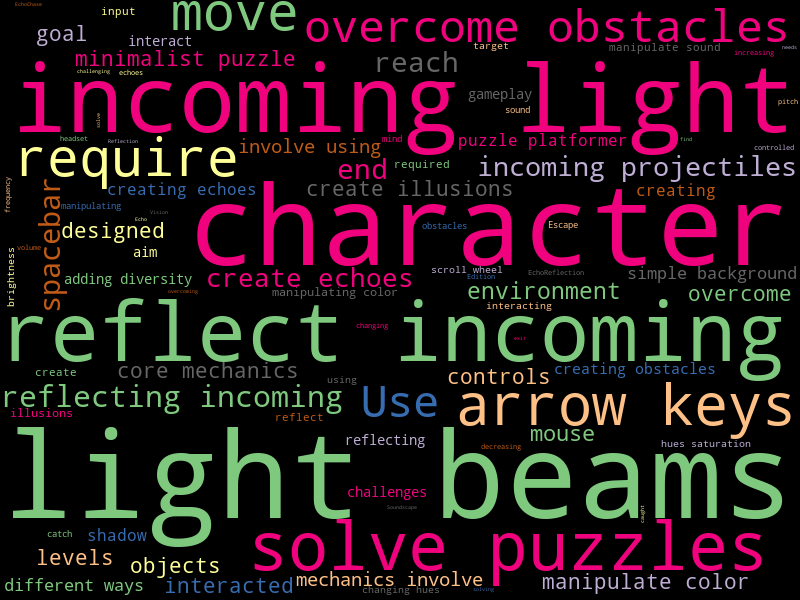}
		}
		
		& 
		\parbox{.45\textwidth}{
			\includegraphics[width=.45\textwidth]{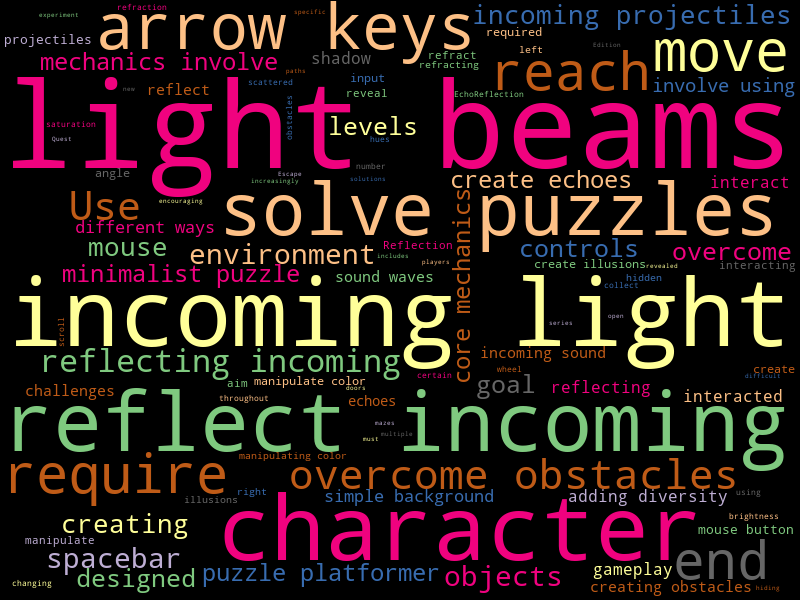}
		} \\
		
		(e) & (f) \\
	\end{tabular}
	\caption{Sequence of word clouds for the video game design task starting from 
		(a) the initial population to the (f) the final population.}
	\label{fig:videogame_cloud2}	
\end{figure*}
 
\end{document}